\documentclass[letterpaper]{article} 
\usepackage{aaai2026}  
\usepackage{times}  
\usepackage{helvet}  
\usepackage{courier}  
\usepackage[hyphens]{url}  
\usepackage{graphicx} 
\usepackage{natbib}  
\usepackage{caption} 
\usepackage{algorithm}
\usepackage{algorithmic}

\usepackage{newfloat}
\usepackage{listings}
\DeclareCaptionStyle{ruled}{labelfont=normalfont,labelsep=colon,strut=off} 
\floatstyle{ruled}
\newfloat{listing}{tb}{lst}{}
\floatname{listing}{Listing}
\newcommand{\metaArgu}{\mathcal{Q}}
\newcommand{\nbset}{\mathbf{
g}}

\nocopyright 

\title{Argumentative Debates for Transparent Bias Detection}
\author {
    Hamed Ayoobi \textsuperscript{\rm 1},
    Nico Potyka\textsuperscript{\rm 2},
    Anna Rapberger\textsuperscript{\rm 3,4},
    Francesca Toni\textsuperscript{\rm 4}
}
\affiliations {
    \textsuperscript{\rm 1}University of Groningen, Netherlands\\
    \textsuperscript{\rm 2}Cardiff University, United Kingdom\\
    \textsuperscript{\rm 3}Technical University Dortmund, Germany\\
    \textsuperscript{\rm 4}Imperial College London, United Kingdom\\
    h.ayoobi@umcg.nl, potykan@cardiff.ac.uk, anna.rapberger@tu-dortmund.de, f.toni@imperial.ac.uk 
}

\usepackage{bibentry}

\usepackage{nicefrac}       
\usepackage{microtype}      
\usepackage{xcolor}         
\usepackage{multirow}
\usepackage{subcaption}  
\usepackage{bm}

\usepackage{todonotes}
\usepackage{amsmath}
\usepackage{amssymb}
\usepackage{mathtools}
\usepackage{amsthm}

\newtheorem{proposition}{Proposition}

\theoremstyle{definition}
\newtheorem{definition}{Definition}

\newtheorem{notation}{Notation}
\usepackage{booktabs}

\newcommand{\acronym}{ABIDE}

\newcommand{\cinput}{\ensuremath{\mathbf{x}}}

\newcommand{\bX}{\ensuremath{\mathbf{X}}}
\newcommand{\bv}{\ensuremath{\mathbf{v}}}
\newcommand{\bV}{\ensuremath{\mathbf{V}}}

\newcommand{\classifier}{\ensuremath{\operatorname{c}}}
\newcommand{\dom}{\ensuremath{\mathcal{D}}}

\newcommand{\classes}{\ensuremath{\mathcal{C}}}
\newcommand{\nb}{\ensuremath{\mathcal{N}}}

\newcommand{\advantaged}{\ensuremath{\mathrm{Adv}_g}}
\newcommand{\disadvantaged}{\ensuremath{\mathrm{Disadv}_g}}
\newcommand{\posG}{\ensuremath{\mathrm{Pos}_{=g}}}
\newcommand{\posNotG}{\ensuremath{\mathrm{Pos}_{\neq g}}}

\newcommand{\agg}{\ensuremath{\operatorname{agg}}}
\newcommand{\infl}{\ensuremath{\operatorname{infl}}}
\newcommand{\core}{\ensuremath{\mathit{Core}}}

\newcommand{\qbaf}{\ensuremath{\mathcal{Q}}}
\newcommand{\strength}{\ensuremath{\sigma}_\qbaf}

\newcommand{\anna}[2][]{\todo[color=red!70!orange!50!white,#1]{AR: #2}}
\newcommand{\FT}[1]{\textcolor{brown}{#1}}

\usepackage[table]{xcolor}
\usepackage{booktabs}

\urldef\mygiturl\url{https://github.com/hamed-ayoobi/ABIDE}

\begin{document}

\maketitle

\begin{abstract}
As the use of AI in society grows, addressing emerging biases is essential to prevent systematic discrimination. Several bias detection methods have been proposed, but, with few exceptions, these tend to ignore transparency. Instead, interpretability and explainability are core requirements for algorithmic fairness, even more so than for other algorithmic solutions, given the human-oriented nature of fairness. We present \acronym\ (Argumentative BIas detection by DEbate), a novel framework that structures \emph{bias detection} transparently as debate, guided by an underlying argument graph as understood in (formal and computational) \emph{argumentation}. The arguments are about the success chances of groups in local \emph{neighbourhoods} and the significance of these neighbourhoods. We evaluate \acronym\ experimentally and demonstrate its strengths in performance against an argumentative baseline. 
\end{abstract}

\section{Introduction}

As the use of AI in society grows, addressing its potential unfairness is becoming increasingly crucial. 
For example, an unfair AI model that supports decision-making in healthcare, discriminating against a certain sub-population, would be extremely harmful. 
In particular, it is essential to address any potential biases against 
specific groups/individuals in data and AI models trained thereupon~\cite{survey-bias21,survey-bias24}.
Various notions of fairness have been proposed in the literature~\cite{survey-bias21,survey-bias24,oana-survey24}. Amongst these notions,  \emph{statistical (or demographic) parity} \cite{stat-parity17} defines fairness as an equal probability of being classified with the desirable (positive) label in different groups. 

Existing solutions to identify unfairness tend to focus on optimising fairness and ignore the need for transparency for the roots of (un)fairness 
\cite{survey-bias24}.
However, transparency, afforded by interpretability and explainability, is crucial for the trustworthy detection and mitigation of bias.
While considerable efforts have been made towards explainability of machine learning models, e.g. as in \cite{ AyoobiSparx, AyoobiFacct}, algorithmic fairness has received limited attention (exceptions include \cite{fairness+explanation22,OanaBias}).

We propose \acronym\ (Argumentative Bias Detection), 
a novel transparent method to detect bias through
argumentative \emph{debates} about the presence of bias 
against individuals, based on the values of \emph{protected features} for the individuals and others in their \emph{neighbourhoods}. 
Debates recently emerged as a powerful mechanism to address various challenges in contemporary AI, such as AI safety~\citep{
debate-safety2024},  
and to improve LLM performance~\citep{debateLLMs-2024}.
In much of 
this literature
, debates are unstructured 
and 
inform other entities that decide the debates' outcome.
Thus, while these debates provide a justification for the outcomes, these
outcomes are not faithfully explained by the debates.

%

\begin{figure}[t]
    \centering
    \includegraphics[width=0.85\linewidth]{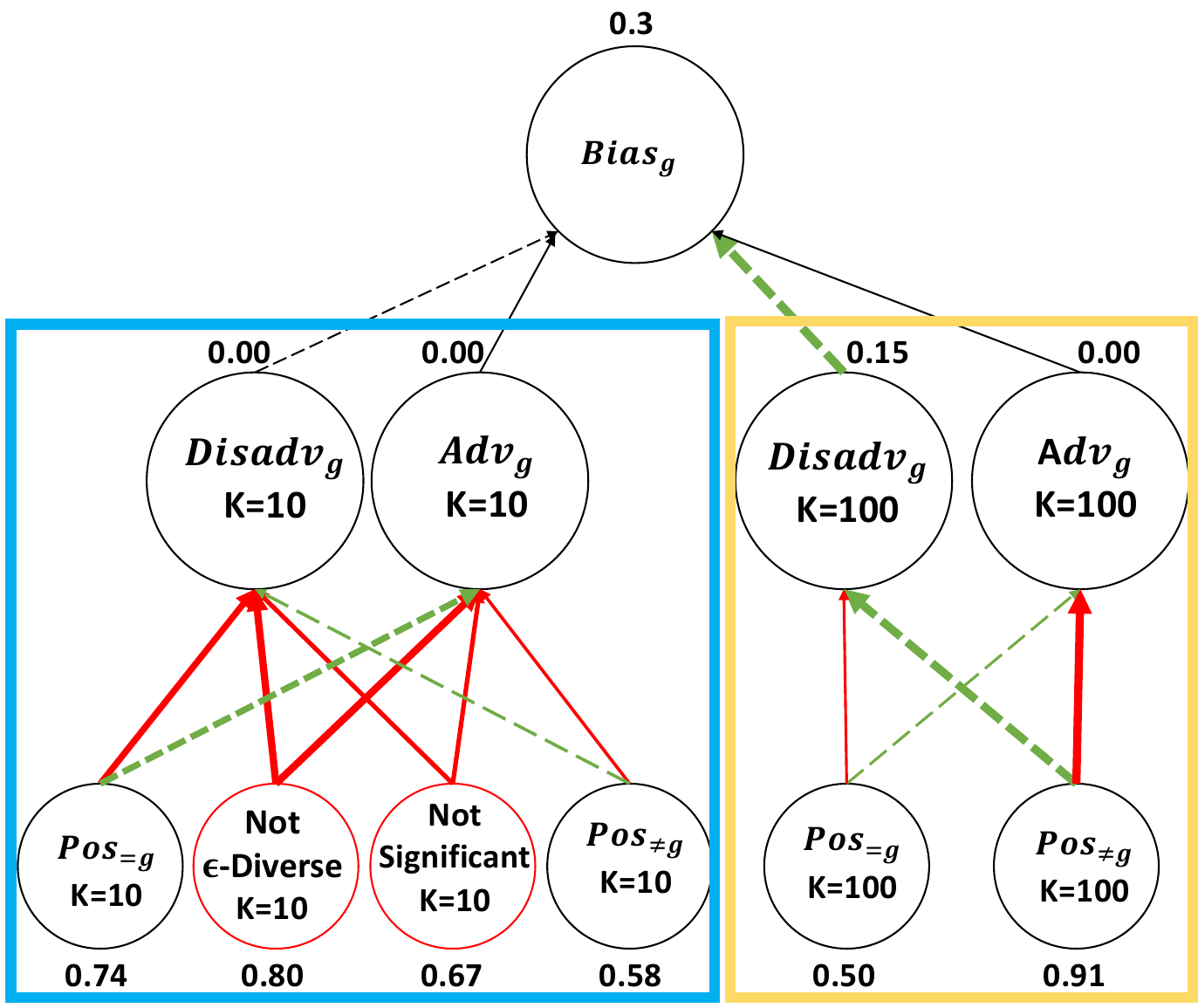}
    \caption{A QBAF generated by 
    \acronym\ for the COMPAS dataset (with protected feature $X_p\!=\!race$ and protected group $g \!=$ \textit{African-American}) for two neighbourhoods of size $K=10, 100$ (critical question arguments with zero strength omitted). Dashed/solid  edges are supports/atttacks. Green/red edges are supports/attacks (from arguments) with nonzero strength, 
    and black edges are 
    (from arguments) with zero strength.
    Edge width reflects the strength of originating arguments
    . (Strengths below/above nodes.
    Details in Section~\ref{sec:main}.)
     }
    \label{fig:arg_compas}
\end{figure}

\acronym\ is designed to guide debates about  the presence of bias
. 
%
To structure debates, we rely upon  \emph{argument schemes} (with critical questions)~\cite{waltonbook,waltonhandbook}, from formal argumentation.
To decide the debates' outcomes
,
we compute arguments' strength with  
\emph{gradual semantics}~\cite{BaroniRT18} from
computational argumentation
. 
%
In summary
:

\begin{enumerate}
    \item We propose a neighbourhood-based notion of fairness for individual bias detection, 
    adapting 
     statistical parity
     .
    \item We define novel argument schemes 
    about the presence of bias based on 
    (properties of) neighbourhoods
    .
    \item We develop a mapping of these schemes into Quantitative Bipolar Argumentation Frameworks (QBAFs)~\cite{BaroniRT18}, enabling structured and modular reasoning 
    for transparent bias detection (see Figure~\ref{fig:arg_compas}).
    \item We formally connect properties of gradual semantics for QBAFs to desirable properties of bias detection
    .
    \item We 
    conduct 
    empirical evaluations 
     with synthetically biased models, models trained on real-world datasets
    , and 
    ChatGPT-4o
   , consistently outperforming
    an argumentative baseline~\cite{OanaBias}.
    \item We show how our approach can empower debate-based bias detection in human-agent and multi-agent scenarios. 
\end{enumerate}

The proofs of all results and  
additional material
can be found in the accompanying 
\cite
{ayoobietal2025}. 
The code is available at \mygiturl.

\section{Related Work}

\paragraph{Fairness in Machine Learning.}

Various 
methods for identifying, measuring and mitigating bias have been proposed~\cite{survey-bias21,survey-bias24}, but they often 
neglect transparency~\cite{oana-survey24}. 
One exception is \cite{fairness+explanation22}, which uses 
explanations to unearth and rectify bias, but relies on feature importance, rather than debates, for transparency. 

Bias can be defined in different ways, 
e.g. by comparing the success probability of the protected group to the one of unprotected individuals
or the general population (\emph{statistical parity}), or by comparing the risk of misclassification of the protected to the one of the non-protected group 
(\emph{predictive equality}) \cite{stat-parity17}.
\acronym\ is based on a local notion of statistical parity. That is, we identify neighbourhoods where groups are locally advantaged/disadvantaged
with respect to success probability. The decision whether a bias exists 
is then 
based on the evidence provided by these neighbourhoods in 
a transparent
way.


\paragraph{Debate and AI.} 
Some approaches rely upon debate protocols  between (two) AI models trained as players in zero-sum games,
e.g. 
towards AI safety~\cite{debate-safety2018,debate-safety2024}
or 
for post-hoc explaining image classifiers~\cite{visualdebates24}.
\cite{debate4images-ECAI24} propose a transparent model based on debate protocols between agents drawing arguments in the form of visual features.
\cite{debateLLMs-2024} use debate to resolve disagreement 
between (two) strong LLMs to inform and improve decisions by weaker models (LLMs or humans).
In all these settings, debates are between two pre-determined entities (e.g. models or agents) and they inform decisions made by other entities (e.g. a verifier 
or a weaker model
). Instead, in our approach,  debates may be in the ``mind'' of a single entity or across two or more entities, as our approach focuses on providing guidance for structuring and formally evaluating (with gradual semantics) debates. 

\paragraph{Argumentation and Bias.}
\cite{waltonbook,waltonhandbook} propose
an argument scheme to capture \emph{arguments from bias}\footnote{
\url{www.rationaleonline.com/map/9gy9gd/argument-from-bias}
},
allowing to draw the conclusion that an arguer is unlikely to have taken both sides of an issue into account if the arguer is biased, subject to addressing critical questions regarding (i) evidence for the arguer's bias and (ii) the need to take multiple sides for that issue. 
Our 
new argument scheme below is orthogonal to this and can be used in combination with it, by
addressing 
the first critical question using properties of neighbourhoods to collect evidence for the bias of the arguer (
a classifier
). 

\citep{OanaBias} were the first to use gradual semantics for 
QBAFs to characterise bias
. While we were inspired by them, our approach differs in 
a simpler QBAF structure (with fewer nodes and relations),
the use of neighbourhoods and their properties (for more robust reasoning on bias), and the empowering of formal links between properties of semantics and of bias detection.

\paragraph{Argumentation and Debate.}
Our approach follows the general idea of using argumentation towards explainability, e.g. as in \cite{AyoobiCVV19, AyoobiCVV21, AyoobiABL, ayoobiICRA, AyoobiPT25, Dejl2025}.
Several 
have pointed to the connection between argumentation and debate, e.g.~\cite{DBLP:journals/argcom/PanissonMB21,maudet} in general, 
and \cite{DBLP:journals/ai/RagoCOT25}
for product recommendation
. While we can leverage on these works towards debate generation, 
our approach is 
orthogonal to them.

\section{Preliminaries}


\paragraph{Classification Problems.} We consider 
variables (features) $X_1, \dots, X_k$ with associated domains
$D_1, \dots, D_k$ and a class variable $Y$ with associated
domain $\classes$ of class labels.
We use $X_p$ to denote a \emph{protected feature} of interest and $g$ to denote its value for the protected group. Thus, 
$X_p \!= \!g$ identifies the protected group and $X_p\!\neq \!g$ the remaining individuals.

For $\dom \!=\! \bigtimes_{i=1}^k D_i$, vectors $\cinput \in \dom$ 
are the \emph{inputs}  of the
classification problem.
%
We consider \emph{(binary) classifiers} $\classifier: \dom \rightarrow \classes\!=\!\{0, 1\}$, 
and regard $1$ as a desirable 
(
e.g., acceptance of a loan application
)
and $0$ as a negative outcome.
%

\begin{notation}
Capital letters refer to variables, 
lowercase letters to values of variables.
Bold capital letters refer to sequences of variables,
bold 
lowercase letters to variable assignments.
For example, if $\bV \!=\! (X_1, X_2, X_3)$, then $\bv$ refers to an assignment $(x_1, x_2, x_3)$ with $x_i \!\in\! D_i$ for $i\!\in\! \{1,2,3\}$.
For each 
$\cinput \in \dom$ and  sequence of variables $\bV = (X_{i_1}, \dots, X_{i_m})$, we let $\cinput|_\bV$ denote the projection of
$\cinput$ onto $\bV$. For example, if $\cinput \!=\! (a,b,c,d)$ and 
$\bV \!=\! (X_2, X_4)$, then $\cinput|_\bV \!=\! (b,d)$.
Finally, we let $\bX$ denote the sequence $(X_1, \ldots, X_k)$ of all variables. 

\end{notation}

\paragraph{
QBAFs.}

Quantitative bipolar argumentation frameworks (QBAFs)~\cite{BaroniRT18}
\label{def:qbaf}
are quadruples $\qbaf =
(\mathcal{A}, \mathcal{R}^{-}, \mathcal{R}^{+}, \tau 
)$ consisting of a finite set of \emph{arguments} $\mathcal{A}$, disjoint binary relations of \emph{attack} $\mathcal{R}^- \!\subseteq\! \mathcal{A} \!\times \!\mathcal{A}$ and \emph{support} $\mathcal{R}^+ \!\subseteq \!\mathcal{A} \!\times\! \mathcal{A}$ 
and a \emph{base score function} $\tau:\mathcal{A}\! \rightarrow \! [0,1]$.
QBAFs can be seen as graphs with arguments 
as nodes and elements of the attack and support relations as edges.
For 
$a\!\in\! \mathcal{A}$, $a^{att}=\{b\in A\mid (b,a)\!\in\! \mathcal{R}^-\}$ and $a^{sup}=\{b\!\in\! A\mid (b,a)\in \mathcal{R}^+\}$ denote the set of all \emph{attackers} and \emph{supporters}, resp.,  of $a$.

To assess the 
acceptability of arguments in a QBAF $\qbaf$, (gradual) semantics can be given  by means of a \emph{strength function} $\strength:\! \mathcal{A} \!\rightarrow \! [0,1]$, 
often defined by an iterative procedure that initializes strength values with the base scores and 
then repeatedly 
updates
the strength of arguments based on the strength
of their 
attackers and supporters. 
For acyclic
QBAFs, this procedure 
always converges and is equivalent to
a linear-time algorithm that first computes a topological ordering of the arguments and then 
updates each argument only once following the order \cite[Proposition 3.1]{Potyka:2019}.

A QBAF semantics is called \emph{modular} if the update function 
can be decomposed into an \emph{aggregation function} $\agg(A ,S)$ that aggregates the strength values $A$ of attackers and $S$ of supporters, 
and an \emph{influence function} $\infl(b, a)$  that adapts the base score $b$ based on the aggregate $a$ \cite{mossakowski2018modular}.
Most QBAF semantics are modular. Examples include the DF-QuAD~\cite{rago2016discontinuity}, Euler-based \cite{amgoud2017evaluation} and
quadratic energy \cite{potyka2018continuous} semantics.
Aggregation functions include \emph{product} $\agg(A ,S) \!=\! \prod_{a \in A} (1 - a) - \prod_{s \in S} (1 - s)$ used in DF-QuAD and \emph{sum}
$\agg(A ,S) \!=\! \sum_{s \in S} s - \sum_{a \in A} a$ used in Euler-based and quadratic energy semantics. 
The influence functions of DF-QuAD and quadratic energy 
both take the form $\infl(b, a) \!=\! b \!-\! b  \cdot  \!f(-s) \!+ \!(1 \!- b) \!\cdot f(s)$, where
$f(x) \!=\! \max \{0,x\}$ for DF-QuAD and $f(x) \!=\! \frac{\max \{0,x\}}{1 + \max \{0,x\}}$ for quadratic energy
.

\paragraph{QBAF Properties.} 
QBAFs 
are often compared 
on 
properties \cite{LeiteM11,AmgoudB16,amgoud2017evaluation, BaroniRT18}
, many of which are tied to properties of aggregation  and influence  functions
\cite{mossakowski2018modular}.
We will use properties of \emph{(strict) monotonicity} and \emph{balance} of these functions from \cite{PotykaB24} 
and recap them in 
\cite
{ayoobietal2025}.
As shown in \cite{PotykaB024}, 
product and sum aggregation 
satisfy balance and monotonicity, sum satisfies strict monotonicity; the influence functions of DF-QuAD and quadratic energy 
satisfy all 
properties.

\section{Neighbourhoods and their Properties}
\label{sec:neigh}

In this section, we will formalize the idea of \emph{local bias}. 
Intuitively, a local bias against a
reference individual exists if other similar
individuals are advantaged by the 
classifier. Formally, these similar
individuals can be 
represented by a
\emph{neighbourhood} 
surrounding the
reference individual.
\begin{definition}
    A \emph{neighbourhood} is a set $\nb \subseteq \dom$. A \emph{neighbourhood 
    of a point} $\cinput \in \dom$ is a 
    set $\nb$ such that $\cinput \in \nb$.
\end{definition}
If we were allowed to 
choose
neighbourhoods  arbitrarily, 
we 
may always 
find one that indicates 
presence
and one that indicates 
absence of 
bias. We thus 
present some
desirable properties 
of neighbourhoods.
%
To begin with, a 
neighbourhood that
contains only few individuals may not be
reliable 
as they 
may 
be exceptions. The larger the
neighbourhood, the larger our confidence
that it is representative of the domain.%
\begin{definition}
Let $N \in \mathbb{N}$.
A neighbourhood $\nb\subseteq \dom$ is called \emph{$N$-significant} if $|\nb| \geq N$.
\end{definition}
Without 
further restrictions,
an adversarial agent could define a neighbourhood by picking 
individuals systematically to demonstrate
the (non-)existence of a bias
. To prevent this
, one 
reasonable assumption is that when we pick
two individuals from a sample, we cannot leave
out individuals between them. Mathematically,
\emph{betweenness} can be described by \emph{convexity}. 
\begin{definition}
Let $S \!\subseteq \!\dom$ be a sample
.
A neighbourhood $\nb\!\subseteq\! S$ is called 
\emph{$S$-objective} if $\cinput_1, \cinput_2 \!\in \!\nb$ and there is an $\cinput_3 \!\in S$ that is a convex combination of $\cinput_1$
, $\cinput_2$, then $\cinput_3 \!\in\! \nb$.
\end{definition}
Formally, an input $\cinput \in \mathbb{R}^n$ is a convex combination of $\cinput_1, \cinput_2  \in \mathbb{R}^n$ if
$\cinput = \lambda \cdot \cinput_1 + (1-\lambda) \cdot \cinput_2$ for some $\lambda \in [0,1]$. Discrete ordinal features such
as Boolean (\{0,1\}) or qualitative descriptions like small (0), medium (1), large (2) can be mapped to integers to match 
the definition. For nominal features such as color or marital status, we demand that they are equal in $\cinput_1$ and $\cinput_2$
(and thus in $\cinput$).

The next proposition explains that $S$-objectivity is always satisfied when we use an \emph{$\epsilon$-neighbourhood}
$\nb_\cinput = \{\cinput' \in S \ \mid \ \|\cinput - \cinput'\| \leq \epsilon \}$ of a point $\cinput$.
It can be defined 
w.r.t. any common distance measure
such as
Euclidean
, Manhattan 
and the Hamming distance
and their weighted variants.
\begin{proposition}
\label{prop:neighbour}
If $\nb_\cinput$ is 
the $\epsilon$-neighbourhood 
of a point $\cinput \in S$ 
with respect to the distance induced by a seminorm\footnote{A seminorm is a function $\| . \|: \dom \rightarrow \dom$ that satisfies sub-additivity (triangle inequality) and absolute homogeneity.
} $\| . \|$, then $\nb$ is $S$-objective.
\end{proposition}

For the remaining properties, we introduce some notation.

\begin{definition}
Given a finite neighbourhood $\nb$ and a partial feature assignment 
$\bv$ to a sequence of variables $\bV$, the \emph{local probability of $\bv$ in 
$\nb$} is
$P_{\nb}(\bv) \!=\! \frac{|\{\cinput \in \nb \ \mid \  (\cinput|_\bV)=\bv\}|}{|\nb|}$.
%
The \emph{local success probability} 
for $\bv$ is defined as
$P_{\nb}(c(\bX)=1 \mid \bv) = \frac{|\{\cinput \in \nb \ \mid \ \cinput|_\bV=\bv, c(\cinput) = 1\}|}{|\{\cinput \in \nb \ \mid \ \cinput|_\bV=\bv\}|}$.   
\end{definition}

    
    


Ideally, our neighbourhoods contain good
representations of both the protected and
non-protected groups. 
For example,
even for a strongly biased model, we 
may be able to find one protected individual
that is treated similar to non-protected ones 
even though the model strongly
discriminates against the non-protected individuals in 
most cases.
Similarly, if it contained
only one non-protected individual, this may
simply be an outlier
.
Since protected groups are often underrepresented, we should not expect  
balanced neighbourhoods. However, if almost
all elements in a neighbourhood belong to
one group, the neighbourhood may not provide very strong
evidence. To quantify 
neighbourhood diversity, we 
use 
entropy.

\begin{definition}
The
\emph{entropy} $H_{\nb}(X_p)$ of the protected feature $X_p$ in $\nb$ is
$H_{\nb}(X_p) \!=\!  - P_{\nb}(X_p\!=\!g) \cdot \log P_{\nb}(X_p\!=\!g) 
    - P_{\nb}(X_p \neq g)  \cdot \log P_{\nb}(X_p \neq g).$
\end{definition}
The entropy takes its maximum $1$ when 
$\nb$ 
is maximally diverse ($P_{\nb}(X_p\!=\!g) \!\!=\!\! P_{\nb}(X_p\! \neq \!g) \!=\! 0.5)$, its 
minimum $0$ when it is minimally 
so
($P_{\nb}(X_p\!=\!g) \!\!=\!\!1$ or $P_{\nb}(X_p\! \neq \! g) \!\!=\!\! 1$),
and 
is monotonically decreasing
in between. 
\begin{definition}
Let $\epsilon \in [0,1]$.
A neighbourhood $\nb\subseteq \dom$ is called 
\emph{$\epsilon$-diverse} if $H_{\nb}(g) = \epsilon$
. 
 \end{definition}
Finally, we 
define 
that a neighbourhood is biased 
if the local conditional
probability of the positive outcome
is significantly larger for the non-protected
than the protected group.
\begin{definition}\label{def:EBA}
A neighbourhood $\nb\subseteq \dom$ is called
 \emph{$\epsilon$-biased against $X_p = g$} if $P_{\nb}(c(\bX)=1 \mid X_p \neq g) - P_{\nb}(c(\bX)=1 \mid X_p = g) = \epsilon$ and $\epsilon >0$.   
\end{definition}

\section{Argument Schemes for Detecting Local Bias}
\label{sec:Argument schemes}

\acronym\ makes use of properties of neighbourhoods to drive argumentative debates about 
the presence of bias against individuals with $X_p=g$. To shape these debates, we define novel argument schemes with accompanying critical questions.
In the next section, we will turn 
both 
into QBAFs
.

%

Figure~\ref{fig:AS for single neighbourhood} gives
our argument scheme  for detecting local bias 
against the protected feature taking the value of interest ($X_p=g$).
%
%
The critical questions address the overall quality (properties) of the neighbourhood
:
\textit{\textbf {CQ1. Is $\mathcal{N}$ $N$-significant?
CQ2. Is $\mathcal{N}$ $S$-objective?
CQ3. Is $\mathcal{N}$ $\epsilon$-diverse?}}
%
When the critical questions are instantiated they give attacks against the conclusion of the (instantiated) argument scheme.

\begin{figure}[t]
    \centering
\begin{tabular}{|ll|}
\hline
{\bf Major premise} & \hspace*{-0.4cm} Generally, if, in 
$\mathcal{N}$, \\
& \hspace*{-0.4cm} $X_p \neq g$ leads to a positive decision \\
& \hspace*{-0.4cm} and $X_p = g$ 
leads to a negative decision, \\ 
& \hspace*{-0.4cm} then $X_p = g$ is 
disadvantaged in $\mathcal{N}$\\
& \hspace*{-0.4cm} and $X_p \neq g$ is 
advantaged in $\mathcal{N}$\\
{\bf Minor premise} & \hspace*{-0.4cm} In the case of the chosen $\mathcal{N}$, \\
&
\hspace*{-0.4cm} $X_p = g$ 
leads to a negative 
decision. \\
{\bf Minor premise} & \hspace*{-0.4cm} In the case of a chosen $\mathcal{N}$, \\
&
\hspace*{-0.4cm} $X_p \neq g$ leads to a positive decision. \\
{\bf Conclusion} & \hspace*{-0.4cm} Th
us,  $X_p = g$ is 
disadvantaged in $\mathcal{N}$ \\
& \hspace*{-0.4cm} and $X_p \neq g$ is 
advantaged in $\mathcal{N}$.\\
\hline
\end{tabular}
    \caption{Argument scheme for 
    neighbourhood $\mathcal{N}$. 
    $X_p = g$ identifies the
protected,  potentially disadvantaged group.
    }
    \label{fig:AS for single neighbourhood}
\end{figure}

We can also combine arguments drawn from different neighbourhoods, via the argument scheme in Figure~\ref{fig:AS for multiple neighbourhoods}.
%
For brevity,  we omit to consider critical questions 
for this scheme.
\begin{figure}[t]
    \centering
\begin{tabular}{|ll|}
\hline
{\bf Major premise}  & 
\hspace*{-0.4cm} If 
$X_p \!\!=\!\! g$ is 
disadvantaged in $\mathcal{N}_1, \ldots, \mathcal{N}_m$,\\
& \hspace*{-0.4cm} then there is a 
bias against $X_p = g$.\\
{\bf Minor premise} & \hspace*{-0.4cm} In the case of chosen $\mathcal{N}_1, \ldots, \mathcal{N}_m$, \\
&
\hspace*{-0.4cm} $X_p \!=\! g$ is 
disadvantaged in $\mathcal{N}_1, \ldots, \mathcal{N}_m$. \\
{\bf Conclusion} & \hspace*{-0.4cm} Th
us,  there is a 
bias against $X_p = g$.\\
\hline
\end{tabular}
    \caption{Argument scheme for combining multiple 
    neighbourhoods $\mathcal{N}_1, \ldots, \mathcal{N}_m$
    .
    }
    \label{fig:AS for multiple neighbourhoods}
\end{figure}

\newcommand{\singlebaf}{Q}
\newcommand{\multibaf}{\mathcal{Q}}
\newcommand{\bases}{\tau}
\newcommand{\valofI}{x_I}
\newcommand{\notvalofI}{x_{\mathit{other}}}
\newcommand{\Xparg}{{\textit{bias}_\nb}}

\section{Arguing about Bias with \acronym}
\label{sec:main}

\acronym\ 
empowers debates 
on bias, 
in
 three steps.
%
\begin{enumerate}
    \item 
    Build a \emph{local bias-QBAF} $\singlebaf_{\nb}$ (based on Figure~\ref{fig:AS for single neighbourhood}).
    \item 
    Add critical questions (CQ1-CQ3).
    \item 
    Build a \emph{global bias-QBAF} $\metaArgu_{\nbset}$ (based on Figure~\ref{fig:AS for multiple neighbourhoods}).
\end{enumerate}
%
We assume a fixed modular gradual semantics $\sigma$ with aggregation function $\agg$ 
and influence function $\infl$ and omit the QBAF subscript when applying $\sigma$.

\paragraph{Arguing about Bias in a single Neighbourhood.}\label{sec:biasInSigle}

The \emph{conclusion} of the argument scheme in Figure~\ref{fig:AS for single neighbourhood} amounts to two arguments, $\disadvantaged$/$\advantaged$, that express that individuals with $X_p = g$ are, resp., disadvantaged/advantaged. 
Their base score is set to $0$.
The \emph{minor premises} give rise to two further arguments $\posG$ and $\posNotG$. 
Their base score corresponds to the group's 
local success probability in the neighbourhood.
The \emph{major premise} informs about the connection between these arguments: $\posG$ attacks/supports $\disadvantaged$/$\advantaged$, while
$\posNotG$  supports/attacks $\disadvantaged$/$\advantaged$.
\begin{definition}\label{def:simple QBAF}
The \emph{local bias-QBAF for $g$ w.r.t. neighbourhood $\nb$}
is the QBAF $\singlebaf_{\nb}=(\mathcal{A}, \mathcal{R}^-,\mathcal{R}^+,\bases)$ with 
    
    $\mathcal{A}=\{\disadvantaged, \advantaged, \posG, \posNotG\}$, 
    
    $\mathcal{R}^-=\{(\posG,\disadvantaged), 
    (\posNotG,\advantaged)\}$, 
    
    $\mathcal{R}^+=
    \{(\posG,\advantaged), (\posNotG,\disadvantaged)\}$,
and 

$\bases(a)\!=\!
\begin{cases}
        P_{\nb}(c(\bX)=1 \mid X_p \neq g),
   & a=\posNotG \\
    P_{\nb}(c(\bX)=1 \mid X_p=g),
   & a=\posG.\\
    0, 
       & else.
\end{cases}
$
\end{definition}
Figure \ref{fig:arg_compas} shows this QBAF (
yellow, right box). 
Intuitively, 
the base score of 
$\disadvantaged$/$\advantaged$ is 0
as our default assumption is that 
individuals with
$X_p\!=\!g$ are not disadvantaged/advantaged.
The base scores of
$\posG$/$\posNotG$ 
are the local success probabilities 
of the groups
$X_p\!=\!g$/ $X_p \!\neq\! g$. 

We show that our choice of QBAF is sensible when choosing the semantics appropriately
.
If $P_{\nb}(c(\bX)\!=\!1 \!\mid\! X_p \!\neq \!g)$ 
and $ P_{\nb}(c(\bX)\!=\!1 \!\mid\! X_p \!=\! g)$ are equal, then there is neither reason to accept
that individuals with $X_p=g$ are disadvantaged nor that they are advantaged. However,
as $P_{\nb}(c(\bX)=1 \mid X_p \neq g)$ becomes larger than $ P_{\nb}(c(\bX)=1 \mid X_p = g)$,
we should gradually accept 
\disadvantaged. Conversely, as $P_{\nb}(c(\bX)=1 \mid X_p \neq g)$ becomes
smaller than $ P_{\nb}(c(\bX)=1 \mid X_p = g)$,
we should gradually accept 
\advantaged.
We can guarantee this behaviour by choosing 
aggregation and influence functions 
with certain properties.
\begin{proposition}
\label{prop_local_impact}
If 
\agg\ and \infl\ satisfy 
\vspace*{-0.1cm}
\begin{enumerate}
    \item monotonicity, then $\sigma(\advantaged) \!\!= \!\!0$ or 
    $\sigma(\disadvantaged) \!\!=\!\! 0$,
    \item balance, then if $P_{\nb}(c(\bX)=1 \mid X_p \neq g) = P_{\nb}(c(\bX)=1 \mid X_p=g)$,
    then $\sigma(\advantaged) = \sigma(\disadvantaged) = 0$,
    \item strict monotonicity, then if $P_{\nb}(c(\bX)=1 \mid X_p \neq g) > P_{\nb}(c(\bX)=1 \mid X_p=g)$,
    then $\sigma(\disadvantaged) > 0$,
    \item  strict monotonicity, then if $P_{\nb}(c(\bX)=1 \mid X_p \neq g) < P_{\nb}(c(\bX)=1 \mid X_p=g)$,
    then $\sigma(\advantaged) > 0$.
\end{enumerate}
\end{proposition}




Item 1 states that we always fully reject at least one of 
$\advantaged$ and $\disadvantaged$.
We fully reject both if both groups have the same local success
 probability
(item 2).
Items 3 and 4 explain that if the protected/non-protected group is at a
disadvantage, the strength of $\disadvantaged$/$\advantaged$ will be
above $0$.

As the inequality
between the protected and non-protected groups increases, the strength
of $\disadvantaged$/$\advantaged$ should change accordingly.
To study this situation, we  compare the strength values 
in two independent local bias-QBAFs, 
connecting  again properties of aggregation and influence functions with the
expected behaviour:

\begin{proposition}
\label{prop:relativeimpact}
Let $\qbaf$ be a QBAF composed of two independent local bias-QBAFs 
$\singlebaf_{\nb^1}, \singlebaf_{\nb^2}$
If 
\agg\ and \infl\
satisfy 
\begin{enumerate}
   \item balance, then, if 
   $P_{\nb^1}(c(\bX)\!=\!1 \mid X_p \!\neq\! g) = P_{\nb^2}(c(\bX)\!=\!1 \mid X_p \!\neq\! g)$
   and $P_{\nb^1}(c(\bX)\!=\!1 \mid X_p \!=\! g) \!=\! P_{\nb^2}(c(\bX)\!=\!1 \mid X_p \!=\! g)$,
   we have $\sigma(\advantaged^1) = \sigma(\advantaged^2)$ and $\sigma(\disadvantaged^1) = \sigma(\disadvantaged^2)$;
   
   \item monotonicity, then if 
   $P_{\nb^1}(c(\bX)=1 \mid X_p \neq g) \geq P_{\nb^2}(c(\bX)=1 \mid X_p \neq g)$
   and $P_{\nb^1}(c(\bX)=1 \mid X_p = g) \leq P_{\nb^2}(c(\bX)=1 \mid X_p = g)$,
  we have $\sigma(\advantaged^1) \leq \sigma(\advantaged^2)$ and $\sigma(\disadvantaged^1) \geq \sigma(\disadvantaged^2)$;
  \item strict monotonicity, then 
  the guarantees of item 2 hold, and, if additionally
  $P_{\nb^1}(c(\bX)=1 \mid X_p \neq g) > P_{\nb^2}(c(\bX)=1 \mid X_p \neq g)$
   or $P_{\nb^1}(c(\bX)=1 \mid X_p = g) < P_{\nb^2}(c(\bX)=1 \mid X_p = g)$,
   then $\sigma(\advantaged^1) < \sigma(\advantaged^2)$ and $\sigma(\disadvantaged^1) > \sigma(\disadvantaged^2)$;
   
   \item monotonicity, then if 
   $P_{\nb^1}(c(\bX)=1 \mid X_p \neq g) \leq P_{\nb^2}(c(\bX)=1 \mid X_p \neq g)$
   and $P_{\nb^1}(c(\bX)=1 \mid X_p = g) \geq P_{\nb^2}(c(\bX)=1 \mid X_p = g)$,
  we have $\sigma(\advantaged^1) \geq \sigma(\advantaged^2)$ and $\sigma(\disadvantaged^1) \leq \sigma(\disadvantaged^2)$;
  
  \item strict monotonicity, then the 
  guarantees of item 4 hold, and, if additionally
  $P_{\nb^1}(c(\bX)=1 \mid X_p \neq g) < P_{\nb^2}(c(\bX)=1 \mid X_p \neq g)$
   or $P_{\nb^1}(c(\bX)=1 \mid X_p = g) > P_{\nb^2}(c(\bX)=1 \mid X_p = g)$,
   then $\sigma(\advantaged^1) > \sigma(\advantaged^2)$ and $\sigma(\disadvantaged^1) < \sigma(\disadvantaged^2)$.
\end{enumerate}
\end{proposition}



Item 1 states that if we find the same inequality in two 
neighbourhoods, we will also have the same strengths
. 
Items 2, 3 deal with the case where the protected group is disadvantaged. Item 2 is a generalization of item 1 and guarantees that
if the non-protected/protected group in $\nb_1$ is at least/most as successful as the non-protected/protected group in $\nb_2$, then 
the strength of $\advantaged$/ $\disadvantaged$ in $\nb_1$ will be at least/most as large as 
in $\nb_2$.
Item 3 guarantees that there will be a strict difference in the strength values if there is a strict difference between the 
probabilities.
Items 4, 5 give symmetrical guarantees 
when the protected group is advantaged.

By design, we 
treat evidence for being advantaged and disadvantaged equally
. 
That is, 
in two neighbourhoods with
symmetric evidence
, the strength of 
$\advantaged$ in one will 
be 
the
strength of 
$\disadvantaged$ in the other:
\begin{proposition}
\label{prop_equity}
For $\singlebaf_{\nb^1}, \singlebaf_{\nb^2}$ as in Prop.~\ref{prop:relativeimpact}, if $P_{\nb^1}(c(\bX)\!=\!1 \!\mid\! X_p\!\neq\! g) \!=\! P_{\nb^2}(c(\bX)\!=\!1 \!\mid\! X_p \!=\! g)$ and
$P_{\nb^1}(c(\bX)\!=\!1 \!\mid\! X_p \!=\! g) \!=\! P_{\nb^2}(c(\bX)\!=\!1 \!\mid \!X_p \!\neq\! g)$,
then
$\sigma(\advantaged^1) \!=\! \sigma(\disadvantaged^2)$ and $\sigma(\advantaged^1) \!=\! \sigma(\disadvantaged^2)$.   
\end{proposition}
Finally, under DF-QuAD, the strength of the $\advantaged$ and $\disadvantaged$ arguments has a particularly natural meaning
:
\begin{proposition}
\label{prop:guarantees}
For $\sigma$ the DF-QuAD semantics, we have 
\vspace*{-0.1cm}
\begin{enumerate}
    \item $\sigma(\advantaged) \!=\! \max \{ 0, P_{\nb}(c(\bX)\!=\!1 \!\mid\! X_p \!=\! g) \!-\! P_{\nb}(c(\bX)\!=\!1 \mid X_p \!\neq\! g)\}$,
    
    \item $\sigma(\disadvantaged) = \max \{ 0, P_{\nb}(c(\bX)=1 \mid X_p \neq g) - P_{\nb}(c(\bX)=1 \mid X_p = g)\}$,
    \item $\sigma(\disadvantaged) = \epsilon$ iff 
    $\nb$ is $\epsilon$-biased against $X_p = g$.
\end{enumerate}
\end{proposition}



\newcommand{\criticalBAF}{\mathcal{Q}^c}

\paragraph{Questioning Bias in a single neighbourhood.}
\label{sec:critical}


The significance of the evidence given by a neighbourhood can be debatable. 
To take account of this, 
we introduce additional arguments 
corresponding to the \emph{critical questions}:
%
    $s$(ignificance) for CQ1, $o$(bjectivity) for CQ2 and $d$(iversity) for CQ3.
    By $CQ=\{s,o,d\}$, we denote the set of our novel 
    arguments. 

    Intuitively, if the answer to these questions is negative, i.e., there is legitimate doubt regarding the significance, diversity, or objectivity of 
    $\nb$, then our confidence in detecting bias within 
    $\nb$ should decrease. 
    Instead, if the answer is positive, then we can trust the bias detection outcome in $\nb$. 
    We thus treat 
    the new arguments as attackers of 
    $\advantaged$ and $\disadvantaged$.
    
The new arguments are unattacked and unsupported by other arguments, and thus their strength (for any $\sigma$) will 
be their base score.
%
So, for 
$s$, the base score should be monotonically
decreasing w.r.t. the 
size of $\nb$ and eventually reach 0, so that small neighbourhoods provide strong attacks. 
For  
$o$, since objectivity is a binary criterion, an
indicator function is most natural.
For 
$d$, the attack strength should be maximal when
the population consists of only one group, and approach 0 as
the distribution becomes uniform: this
behaviour can be captured by 
the entropy, as given in Section~\ref{sec:neigh}. However, since
we cannot expect a uniform distribution for underrepresented
groups, we should be able to rescale the entropy: our base
score function for $d$ should thus be monotonically
decreasing
.
Formally, critical questions lead to the QBAF below:
    
\begin{definition}\label{def:critical QBAF}
   Let $\singlebaf_{\nb}\!=\!(\mathcal{A}, \mathcal{R}^-,\mathcal{R}^+,\bases)$ be a local bias-QBAF. 
    The corresponding \emph{local bias-QBAF with critical questions} is  
    $\criticalBAF_\nb\!=\!(\mathcal{A}', {\mathcal{R}^-}',\mathcal{R}^+,\bases')$ with 
$\mathcal{A}'\!=\!\mathcal{A}\!\cup\! CQ$,
${\mathcal{R}^-}'\!=\!\mathcal{R}^-\!\cup\! 
 \{(a,\advantaged), (a,\disadvantaged)\!\mid\! a\!\in\! CQ\}$
 and $\bases'$ is such that 
 $\bases'(a)=\bases(a)$ for all $a\not\in CQ$, 
 and
 \begin{itemize}
 \item 
    $\bases'(s) = f^s(|\nb|)$ for a monotonically decreasing function
    $f^s\!:\!\mathbb{N}\!\rightarrow\! [0,1]$ such that  $f^s(1)\!=\!1$, and 
    $\lim_{x \!\rightarrow\! \alpha} f^s(x) \!=\! 0$
    for some \emph{significance threshold} $\alpha\!\in \!\mathbb{N} \!\cup\! \{\infty\}$;
    \item 
    $\bases'(o) = f^o(\nb)$ where 
    $f^o(\nb)\!=\!0$ if $\nb$ is $S$-objective, and $f^o(\nb)\!=\! \gamma$ 
    else, where  $\gamma \in (0,1]$ is an \emph{objectivity weight};
    \item 
    $\bases'(d) = f^d( H_{\nb}(g) )$ where
    for $f^d\!:\![0,1]\!\rightarrow \![0,1]$ is
    monotonically decreasing
    and satisfies $f^d(0)\!=\! 1$ and $f^d(\beta)\!=\!0$
    for some \emph{entropy threshold} $\beta \in (0,1]$. \end{itemize}
\end{definition}

We illustrate this QBAF in Figure \ref{fig:arg_compas} (in the blue, 
left box).
In the experiments (Section~\ref{sec:experiments}), we use
$f^s(x)=\frac{\max \{\alpha - x, 0\}}{\alpha}$ ($\alpha \!<\! \infty$)
and $f^d(x) \!=\! 1 \!-\! \max \{\frac{x}{\beta}, 1\}$.

Similarly to Section~\ref{sec:biasInSigle}, 
properties of $\sigma$ guarantee that 
the new arguments behave as intended. 
We do not go into formal detail 
but observe that,
if 
\agg\ and \infl\ satisfy
    balance, then 
    $\advantaged/\disadvantaged$ will remain unaffected by critical questions that do not apply (base score $0$), and,
    if they satisfy monotonicity, then the strength of 
    $\advantaged/\disadvantaged$ will decrease monotonically w.r.t. the strength (i.e. base score) of 
   $s,o,d$.

\paragraph{Arguing about Bias across neighbourhoods.}\label{sec:biasAcrossNeighbourdhoods}

\newcommand{\topicarg}{
{\ensuremath{\mathrm{bias}_g}}}

Finally, we combine multiple local bias-QBAFs 
for different
neighbourhoods $\nb_1,\dots,\nb_m$, 
as per Figure~\ref{fig:AS for multiple neighbourhoods}
. 
Their individual $\advantaged/\disadvantaged$ arguments
attack/support a \emph{global bias} argument $\topicarg$. 
We make the following assumptions
: 
we have no prior information about biases of the classifier and, 
in the absence of information, we 
take that 
the classifier is unbiased. 
We capture this by setting the base score of $\topicarg$ to $0$. 

\begin{definition}\label{def:simple QBAF}
Let $\singlebaf^c_{\nb_i}\!=\!(\mathcal{A}_i, {\mathcal{R}_i}^-, {\mathcal{R}_i}^+,\bases_i)$ be the local bias-QBAF with critical questions for $g$ w.r.t. $\nb_i\!\in \!\{\nb_1,\dots,\nb_m\}$.
Then, the \emph{global bias-QBAF for $g$} is the QBAF $\metaArgu_{\nbset}=(\mathcal{A},\mathcal{R}^-,\mathcal{R}^+,\bases)$ with

\quad\quad  $\mathcal{A}=\{\topicarg\}\cup \bigcup_{1 \leq i\leq m} \mathcal{A}_i$;

\quad\quad  $\mathcal{R}^-=
\{(\advantaged^i,\topicarg)\mid 
1 \leq i \leq m\} \cup \bigcup_{i=1}^m \mathcal{R}_i^-
$;

\quad\quad  $\mathcal{R}^+=\{(\disadvantaged^i,\topicarg)\mid 
1 \leq i \leq m\} \cup \bigcup_{i=1}^m \mathcal{R}_i^+$;

\quad\quad 
$\bases(a)=\bases_i(a)$ if $a \in \mathcal{A}_i$ and 0 if $a=\topicarg$.
\end{definition}
 Figure \ref{fig:arg_compas} illustrates 
 a global bias-QBAF, for $m=2$.

Again, 
properties of $\sigma$ can guarantee that local bias-QBAFs affect 
\topicarg\
as expected. 
In particular, 
\topicarg\
will remain unaffected by fully rejected local bias arguments (strength $0$)
and its strength will increase/decrease monotonically w.r.t. local \disadvantaged/\advantaged\ arguments. 

\begin{proposition}
\label{prop:global}
For  $\metaArgu_{\nbset}\!=\!(\mathcal{A},\mathcal{R}^-,\mathcal{R}^+,\bases)$ 
a global bias-QBAF
, let $S_d \!\!=\!\! \{\disadvantaged^i \mid   
i\!\in\! \{1, \ldots, m\}, \sigma(\disadvantaged^i) \!>\! 0 \}$
and 
$S_a \!=\! \{\advantaged^i \!\mid\! 
i\!\in\! \{1, \ldots, m\}, \sigma(\advantaged^i) \!>\! 0 \}$.
If 
\agg\ and \infl\ satisfy balance and monotonicity, then
 1. $S_d \succeq S_a$ implies $
 \sigma(\topicarg) \geq 0$,
 and   
    2. $S_a \succeq S_d$ implies $
    \sigma(\topicarg) = 0$.
\end{proposition}
Note that item 1 of Proposition \ref{prop_local_impact} implies that $S_d \cap S_a = \emptyset$.





\section{Experiments}\label{sec:experiments}

We conduct 
experiments 
with: 1.  synthetically biased classifiers 
(two hand-crafted 
and one
resulting from training on the Adult Census Income (ACI) dataset~\cite{adult_census}, also used for testing
);  2. trained classifiers  (
logistic regression, but our method 
applies to any classifier%
)
; 3. LLMs (ChatGPT-4o)
. 
For training (2) and for testing (2,3) we 
use the Bank Marketing \cite{bank_marketing} and COMPAS \cite{compas_kaggle, compas_paper}  datasets.
The datasets we choose are
commonly used in the fairness literature \cite{COMPAS_Facct2024, COMPAS_NEURIPS2024, NEURIPS2024_Fairness}. 
%
We consider also LLMs 
as they are prone to societal biases due to their exposure to human-generated data \cite{BiasLLMBreakingDown, BiasGenAILaw}.
In all experiments, we compare \acronym\ 
with the 
argumentative 
approach by \cite{OanaBias} 
(\emph{IRB} in short).  
We report results with  quadratic energy 
(DF-QuAD  
gave similar results
). 
We report results for 1000 randomly sampled individuals and their KNN-neighbourhoods~\cite{KNN_original_paper}, drawn from the test sets of the chosen datasets. A sensitivity analysis evaluating the model's performance across varying parameters is provided in 
\cite
{ayoobietal2025}.

\paragraph{Experiments with Synthetically Biased Models.}
\label{sec:synt}
We developed three synthetic 
models to obtain an objective ground truth in the setting of the ACI dataset: two \emph{globally biased models} (
Global 1 and Global 2) and one \emph{locally biased model} (Local 1). 
Global 1 and 2 are hand-crafted decision trees (see Figures~5, 6 in  Appendix~3):
the former predicts a negative label for all female individuals and a positive label for all others; the latter
 predicts a negative label for all black female individuals.
For Local 1, we trained a logistic regression model on the 
ACI training set. In contrast to the global  models, Local 1 operates at a neighbourhood level, changing the 
prediction to the negative label for female individuals if their neighbourhood exhibits an $\epsilon$-bias against $gender=female$ in the originally trained model (see 
Figure~7 in \cite
{ayoobietal2025}). 

\textit{\textbf{Bias in Single Neighbourhoods.}}
Table~\ref{tab:single_neighbourhood} shows that
our method consistently outperforms IRB with \textit{Global 2} and \textit{Local 1} (IRB yields all zero scores 
with \textit{Global 2}, as it 
focuses on showing bias against individual 
features, rather than combinations thereof). 
For \textit{Global 1}, the two approaches perform similarly.
Overall, our method runs significantly faster due to a simpler QBAF with fewer nodes and relations. 

\begin{table}
\centering
\resizebox{\linewidth}{!}{
\begin{tabular}{lcccccccc}
\toprule
Method & Model & K & Accuracy & Precision & Recall & F1-score & Runtime (s)  \\
\midrule
Our & \multirow{2}{*}{Global 1} & \multirow{2}{*}{50} & 0.95 $\pm$ 0.01 & 1.00 $\pm$ 0.00 & 0.90 $\pm$ 0.02 & 0.95 $\pm$ 0.01 & \textbf{3.87 $\pm$ 0.03} \\
IRB &  &  & 0.95 $\pm$ 0.01 & 1.00 $\pm$ 0.00 & 0.90 $\pm$ 0.01 & 0.95 $\pm$ 0.02 & 28.22 $\pm$ 0.04 \\
\midrule
Our & \multirow{2}{*}{Global 1} & \multirow{2}{*}{100} & 1.00 $\pm$ 0.00 & 1.00 $\pm$ 0.00 & 1.00 $\pm$ 0.00 & 1.00 $\pm$ 0.00 & \textbf{3.88 $\pm$ 0.02} \\
IRB &  &  & 1.00 $\pm$ 0.00 & 1.00 $\pm$ 0.00 & 1.00 $\pm$ 0.00 & 1.00 $\pm$ 0.00 & 48.47 $\pm$ 0.05 \\
\midrule
Our & \multirow{2}{*}{Global 1} & \multirow{2}{*}{200} & 1.00 $\pm$ 0.00 & 1.00 $\pm$ 0.00 & 1.00 $\pm$ 0.00 & 1.00 $\pm$ 0.00 & \textbf{3.82 $\pm$ 0.02} \\
IRB &  &  & 1.00 $\pm$ 0.00 & 1.00 $\pm$ 0.00 & 1.00 $\pm$ 0.00 & 1.00 $\pm$ 0.00 & 97.81 $\pm$ 0.04 \\
\midrule

Our & \multirow{2}{*}{Global 2} & \multirow{2}{*}{50} & \textbf{0.96 $\pm$ 0.02} & \textbf{1.00 $\pm$ 0.01} & \textbf{0.96 $\pm$ 0.02} & \textbf{0.98 $\pm$ 0.01} & \textbf{0.34 $\pm$ 0.02} \\
IRB &  &  & 0.00 $\pm$ 0.00 & 0.00 $\pm$ 0.00 & 0.00 $\pm$ 0.00 & 0.00 $\pm$ 0.00 & 4.28 $\pm$ 0.03 \\
\midrule
Our & \multirow{2}{*}{Global 2} & \multirow{2}{*}{100} & \textbf{1.00 $\pm$ 0.00} & \textbf{1.00 $\pm$ 0.00} & \textbf{1.00 $\pm$ 0.00} & \textbf{1.00 $\pm$ 0.00} & \textbf{0.35 $\pm$ 0.01} \\
IRB &  &  & 0.00 $\pm$ 0.00 & 0.00 $\pm$ 0.00 & 0.00 $\pm$ 0.00 & 0.00 $\pm$ 0.00 & 8.99 $\pm$ 0.04 \\
\midrule
Our & \multirow{2}{*}{Global 2} & \multirow{2}{*}{200} & \textbf{1.00 $\pm$ 0.00} & \textbf{1.00 $\pm$ 0.00} & \textbf{1.00 $\pm$ 0.00} & \textbf{1.00 $\pm$ 0.00} & \textbf{0.38 $\pm$ 0.01} \\
IRB &  &  & 0.00 $\pm$ 0.00 & 0.00 $\pm$ 0.00 & 0.00 $\pm$ 0.00 & 0.00 $\pm$ 0.00 & 17.44 $\pm$ 0.05 \\
\midrule
Our & \multirow{2}{*}{Local 1} & \multirow{2}{*}{50} & \textbf{1.00 $\pm$ 0.00} & \textbf{1.00 $\pm$ 0.00} & \textbf{1.00 $\pm$ 0.00} & \textbf{1.00 $\pm$ 0.00} & \textbf{5.00 $\pm$ 0.05} \\
IRB & &  & 0.81 $\pm$ 0.02 & \textbf{1.00 $\pm$ 0.00} & 0.44 $\pm$ 0.02 & 0.61 $\pm$ 0.02 & 30.66 $\pm$ 0.05 \\
\midrule
Our & \multirow{2}{*}{Local 1} & \multirow{2}{*}{100} & \textbf{1.00 $\pm$ 0.00} & \textbf{1.00 $\pm$ 0.00} & \textbf{1.00 $\pm$ 0.00} & \textbf{1.00 $\pm$ 0.00} & \textbf{4.96 $\pm$ 0.04}\\
IRB &  &  & 0.74 $\pm$ 0.02 & \textbf{1.00 $\pm$ 0.00} & 0.31 $\pm$ 0.02 & 0.48 $\pm$ 0.02 & 49.45 $\pm$ 0.05 \\
\midrule
Our & \multirow{2}{*}{Local 1} & \multirow{2}{*}{200} & \textbf{1.00 $\pm$ 0.00} & \textbf{1.00 $\pm$ 0.00} & \textbf{1.00 $\pm$ 0.00} & \textbf{1.00 $\pm$ 0.00} & \textbf{5.07 $\pm$ 0.03} \\
IRB &  &  & 0.70 $\pm$ 0.01 & \textbf{1.00 $\pm$ 0.00} & 0.22 $\pm$ 0.01 & 0.36 $\pm$ 0.01 & 97.16 $\pm$ 0.04\\
\bottomrule
\end{tabular}
}
\caption{The performance of our method (using $\singlebaf^c_{\nb_i}$) versus IRB, with single neighbourhoods ($\nb_i$) of different sizes (K) for synthetically biased classifiers. Best results in bold.}
\label{tab:single_neighbourhood}
\end{table}


\begin{table}
\centering
\resizebox{\linewidth}{!}{
\begin{tabular}{lccccc}
\toprule
Method  & Model & Accuracy & Precision & Recall & F1-score \\
\midrule
Our  & \multirow{2}{*}{Global 1} & 1.00 $\pm$ 0.00 & 1.00 $\pm$ 0.00 & 1.00 $\pm$ 0.00 & 1.00 $\pm$ 0.00 \\
IRB  &  & 1.00 $\pm$ 0.00 & 1.00 $\pm$ 0.00 & 1.00 $\pm$ 0.00 & 1.00 $\pm$ 0.00 \\
\midrule
Our  & \multirow{2}{*}{Global 2} & \textbf{1.00 $\pm$ 0.00} & \textbf{1.00 $\pm$ 0.00} & \textbf{1.00 $\pm$ 0.00} & \textbf{1.00 $\pm$ 0.00} \\
IRB  &  & 0.00 $\pm$ 0.00 & 0.00 $\pm$ 0.00 & 0.00 $\pm$ 0.00 & 0.00 $\pm$ 0.00 \\
\midrule
Our  & \multirow{2}{*}{Local 1}  & \textbf{1.00 $\pm$ 0.00} & \textbf{1.00 $\pm$ 0.00} & \textbf{1.00 $\pm$ 0.00} & \textbf{1.00 $\pm$ 0.00} \\
IRB  &  & 0.70 $\pm$ 0.02 & 1.00 $\pm$ 0.01 & 0.22 $\pm$ 0.02 & 0.36 $\pm$ 0.01 \\
\bottomrule
\end{tabular}
}
\caption{The performance of our method (using $\metaArgu_{\nbset}$ with $\nb_1,\nb_2, \nb_3$ of sizes K=50,100,200, respectively) versus IRB (with the largest neighbourhood $\nb_3$) for synthetically biased classifiers. Best results in bold.}
\label{tab:across_neighbourhoods}
\end{table}


\textit{\textbf{Bias Across Neighbourhoods.}}
We 
consider multiple 
neighbourhoods of varying sizes (50, 100, 200).\footnote{We 
 run experiments with other 
 sizes, ranging from $K=10$ to $K=200$ in intervals of 10. 
 See \cite
 {ayoobietal2025} for details.} 
 %
Table~\ref{tab:across_neighbourhoods} shows that
our method achieves perfect performance for all classifiers.
IRB fails for Global 2 for the same reasons as before.



\paragraph{Experiments with Trained Models.}
\label{sec:qual}

\noindent 
We trained logistic regression models on Bank marketing and COMPAS
. Here, we no longer have ground truth labels for bias and
thus just compare the number of identified biases
.
%
\begin{table}[h]
\centering
\begin{subtable}[b]{0.48\linewidth}
  \centering
  \resizebox{\linewidth}{!}{%
    \begin{tabular}{|c|l|c|c|} \hline 
 \multicolumn{2}{|c|}{\multirow{2}{*}{\textbf{Feature value}}}& \multicolumn{2}{|c|}{\textbf{Logistic Regression}} \\
      \cline{3-4}
      \multicolumn{2}{|c|}{} & \textbf{IRB} & \textbf{Our}     \\ \hline
      \multirow{6}{*}{\rotatebox[origin=c]{90}{race}} 
      & African-American & 2 & 77   \\ 
      & Caucasian & 1 & 1   \\   
      & Hispanic & 26 & 13   \\ 
      & Asian & 3 & 0   \\ 
      & Native American & 3 & 0   \\ 
      & Other & 11 & 0   \\
      \hline
      \rotatebox[origin=c]{90}{sex}
      & Female & 1 & 0   \\  \hline
    \end{tabular}
  }
  \caption{COMPAS dataset}
  \label{tab:compas}
\end{subtable}
\hspace{0.02\linewidth}
\begin{subtable}[b]{0.48\linewidth}
  \centering
  \resizebox{\linewidth}{!}{%
    \begin{tabular}{|c|l|c|c|} \hline 
 \multicolumn{2}{|c|}{\multirow{2}{*}{\textbf{Feature value}}}& \multicolumn{2}{|c|}{\textbf{Logistic Regression}} \\
      \cline{3-4}
      \multicolumn{2}{|c|}{} & \textbf{IRB} & \textbf{Our}      \\ \hline 
      \multirow{2}{*}{\rotatebox[origin=c]{90}{age}} 
      & MidAge & 0 & 0 \\ 
      & YoungOrOld & 20 & 78 \\ \hline
      \multirow{3}{*}{\rotatebox[origin=c]{90}{marital}} 
      & Married & 0 & 4  \\ 
      & Single & 4 & 11 \\ 
      & Divorced & 19 & 10 \\ \hline
    \end{tabular}
  }
  \caption{Bank Marketing dataset}
  \label{tab:bank}
\end{subtable}%
\caption{Count of feature values being biased against in queried individuals  
using $\metaArgu_{\nbset}$ with 
$\nb_1,\nb_2, \nb_3$ of sizes K=50,100,200, resp., and  IRB (with 
$\nb_3$).
}
\label{tab:combined_comparison}
\end{table}
Table \ref{tab:compas} 
gives results drawn from COMPAS, showing that our method identifies substantially more biased cases against African-American individuals (77 vs. 2),
in line with 
the literature \cite{
compas_paper, COMPAS_AAAI2020}, while detecting fewer or no biased cases for 
the other groups compared to IRB. 
 Table \ref{tab:bank} gives results for 
individuals identified as biased against specific age and marital status groups in the Bank Marketing dataset
. In particular, for ``YoungOrOld'' age (younger than 25 years of age or older than 60 years of age), ``Married'', and ``Single'' marital features, our method detects more biased cases than IRB. 
We show examples in \cite
{ayoobietal2025}.

\paragraph{
Experiments with LLMs.}\label{sec:llm}
 We examine the widely used ChatGPT-4o
.
We followed the same protocol as in \cite{LLM_Bias}, 
prompting the model to predict the class label based on the individuals' feature values.
Table \ref{tab:LLM_comparison} shows the results of our experiments. 
Since the model is not fine-tuned on  COMPAS or Bank Marketing, the results reveal the presence of inherent biases within the model itself
. In Table \ref{tab:compas_LLM}, our method identifies a significantly higher number of biased instances against African Americans (129 cases) compared to zero by IRB
. Similarly, the model demonstrates a higher degree of bias against females (6 instances)
. 
In Table \ref{tab:bank_LLM}, our approach also uncovers bias 
which IRB fails to detect.
These results suggest that our method may be more sensitive in uncovering latent bias patterns that IRB overlooks.

\begin{table}[h]
\centering
\begin{subtable}[b]{0.48\linewidth}
  \centering
  \resizebox{\linewidth}{!}{%
    \begin{tabular}{|c|l|c|c|} \hline 
 \multicolumn{2}{|c|}{\multirow{2}{*}{\textbf{Feature value}}} &\multicolumn{2}{|c|}{\textbf{LLM (GPT4)}}\\
      \cline{3-4}
      \multicolumn{2}{|c|}{}    & \textbf{IRB} &\textbf{Our}   \\ \hline
      \multirow{6}{*}{\rotatebox[origin=c]{90}{race}} 
      & African-American  & 0&129\\ 
      & Caucasian &  0&2\\   
      & Hispanic  & 8&1\\ 
      & Asian   & 2&0\\ 
      & Native American   & 3&0\\ 
      & Other   & 2&1\\
      \hline
      \rotatebox[origin=c]{90}{sex}
      & Female    & 0&6\\  \hline
    \end{tabular}
  }
  \caption{COMPAS dataset}
  \label{tab:compas_LLM}
\end{subtable}
\hspace{0.02\linewidth}
\begin{subtable}[b]{0.48\linewidth}
  \centering
  \resizebox{\linewidth}{!}{%
    \begin{tabular}{|c|l|c|c|} \hline 
 \multicolumn{2}{|c|}{\multirow{2}{*}{\textbf{Feature value}}} & \multicolumn{2}{|c|}{\textbf{LLM (GPT4)}}\\
      \cline{3-4}
      \multicolumn{2}{|c|}{}    & \textbf{IRB} &\textbf{Our}   \\ \hline 
      \multirow{2}{*}{\rotatebox[origin=c]{90}{age}} 
      & MidAge  &0 & 2\\ 
      & YoungOrOld &0&14 \\ \hline
      \multirow{3}{*}{\rotatebox[origin=c]{90}{marital}} 
      & Married  &0&13 \\ 
      & Single &0&18 \\ 
      & Divorced &0&0 \\ \hline
    \end{tabular}
  }
  \caption{Bank Marketing dataset}
  \label{tab:bank_LLM}
\end{subtable}%
\caption{Count of feature values being biased against in queried individuals 
using
$\metaArgu_{\nbset}$ with 
$\nb_1,\nb_2, \nb_3$ of sizes K=50,100,200, resp., versus  IRB (with 
$\nb_3$). 
}
\label{tab:LLM_comparison}
\end{table}


\section{Argumentative Debates} \label{sec: ArgDebateExample}

\begin{figure}
    \centering
    \includegraphics[width=1\linewidth]{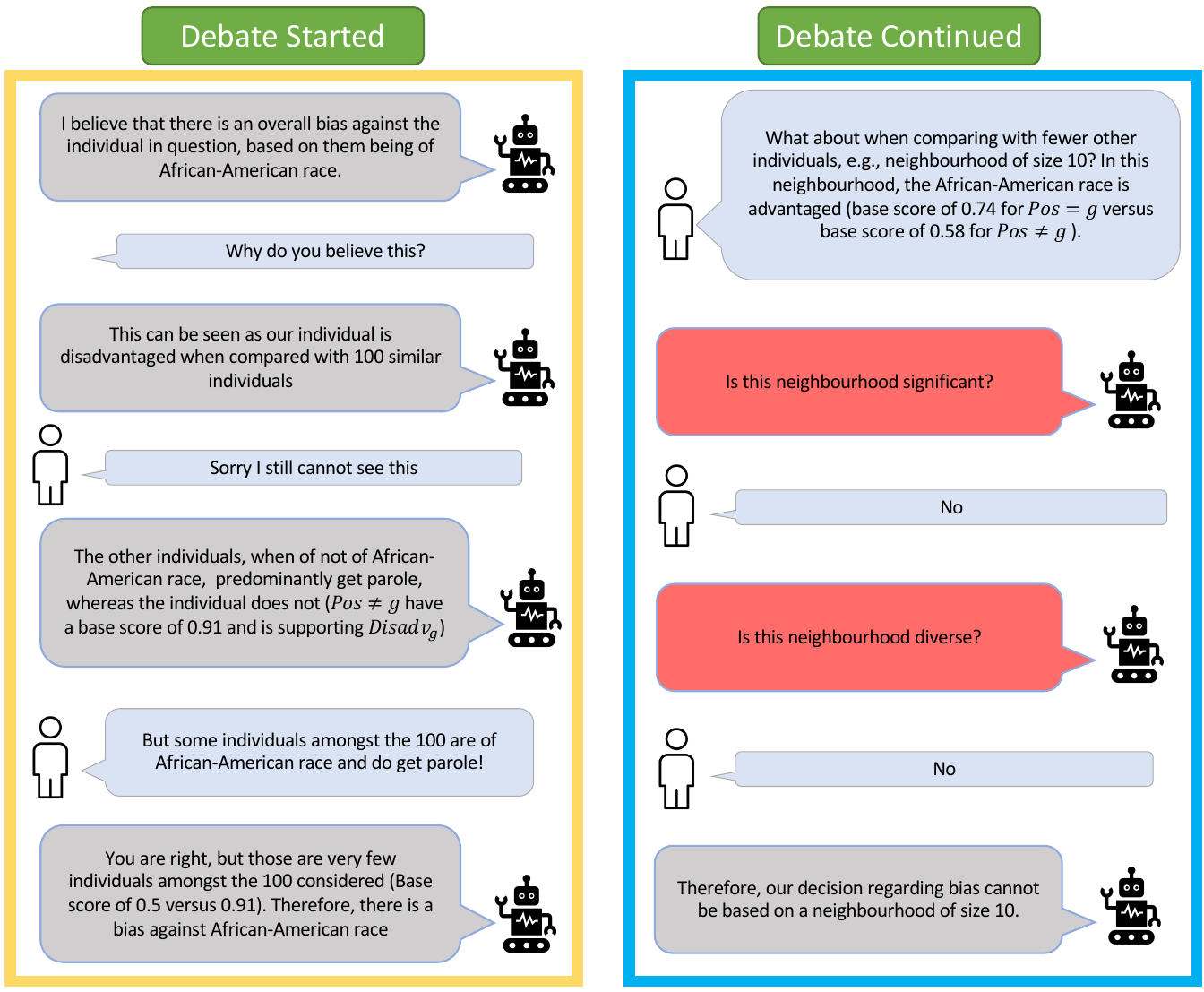}
    \caption{Example of an Argumentative Debate (the debates in the yellow/blue boxes stem from the 
    corresponding boxes in Figure \ref{fig:arg_compas}). }
    \label{fig:debate_visualized}
\end{figure}

Our QBAFs 
can be used 
as debate-based explanations.
The debates are naturally argumentative, by presenting evidence for and against bias, and for and against this evidence.  
These debates may
take place in the ``mind'' of a \emph{single agent}, taking 
two opposing role, of a \emph{proponent} arguing that there is a bias 
and an \emph{opponent} arguing that there is no such bias, in the spirit of dispute trees and dispute derivations~\cite{DBLP:journals/logcom/ThangDH09,DBLP:journals/flap/CyrasFST17}.
As an example,
for the QBAF in Figure~\ref{fig:arg_compas},
the proponent may state
the top most argument in Figure~\ref{fig:debate_visualized} (left) 
and 
the opponent may then continue
by stating: ``But some individuals amongst the 100 are of African-American race and do get parole!'' 
This behaviour can be obtained 
with conversational templates 
as in \cite{DBLP:journals/ai/RagoCOT25}.
%
Also, the QBAFs can 
be used as
templates to drive debates 
amongst \emph{two or more agents}
.
To do so, roles (for and against bias) or specific neighbourhoods could be associated with agents,
with each 
using critical question arguments to criticise the neighbourhoods chosen by the others.
Finally, the QBAFs can 
be used as the backbone 
for debates between \emph{agents and humans} (as in Figure~\ref{fig:debate_visualized}). 
%
When debates arise within or amongst agents, they could be generated and presented as explanations to users in one go, or incrementally, based on the user's prompts (as in
Figure~\ref{fig:debate_visualized}).
This requires the ability of agents to match the prompts with specific parts of the underpinning QBAF, and generate replies based on relevant parts 
thereof.
We leave this to future work.

\section{Conclusion}
We presented \acronym, a novel model-agnostic approach for bias detection 
for analysing both local and global biases through 
argument schemes and quantitative bipolar argumentation. Our approach addresses a critical 
need for algorithmic fairness, by integrating transparency into bias detection, 
%
while also exhibiting desirable properties, 
linked to established properties
for quantitative bipolar argumentation
, and performance advantages over an  argumentative baseline.

We discussed how \acronym\ can support debates about bias. 
Future work will focus on the realisation thereof within and across agents, as well as between agents and humans, including with user studies. 
It will be especially interesting to explore how humans can contribute to the debates, specifically by contesting 
existing arguments and adding new arguments to the debate, in the spirit of~\cite{AyoobiKR}.

\section*{Acknowledgements}
Ayoobi, Rapberger, and Toni were funded by 
the ERC under the EU’s Horizon 2020 research and innovation programme (ADIX, grant number 101020934). Toni was also funded by J.P. Morgan and by the Royal Academy of Engineering under the Research Chairs and Senior Research Fellowships scheme.

\bibliography{aaai2026}

\begin{thebibliography}{2}
\providecommand{\natexlab}[1]{#1}

\bibitem[{Potyka and Booth(2024)}]{PotykaB24}
Potyka, N.; and Booth, R. 2024.
\newblock An Empirical Study of Quantitative Bipolar Argumentation Frameworks for Truth Discovery.
\newblock In \emph{COMMA 2024}, 205--216. {IOS} Press.

\bibitem[{Saltelli et~al.(2008)Saltelli, Ratto, Andres, Campolongo, Cariboni, Gatelli, Saisana, and Tarantola}]{Saltelli2008}
Saltelli, A.; Ratto, M.; Andres, T.; Campolongo, F.; Cariboni, J.; Gatelli, D.; Saisana, M.; and Tarantola, S. 2008.
\newblock \emph{Global Sensitivity Analysis: The Primer}.
\newblock Chichester, UK: John Wiley \& Sons.
\newblock ISBN 978-0-470-05997-5.

\end{thebibliography}


\begin{thebibliography}{50}
\providecommand{\natexlab}[1]{#1}

\bibitem[{Amgoud and Ben{-}Naim(2016)}]{AmgoudB16}
Amgoud, L.; and Ben{-}Naim, J. 2016.
\newblock Axiomatic Foundations of Acceptability Semantics.
\newblock In \emph{{KR} 2016}, 2--11. {AAAI} Press.

\bibitem[{Amgoud and Ben-Naim(2017)}]{amgoud2017evaluation}
Amgoud, L.; and Ben-Naim, J. 2017.
\newblock Evaluation of arguments in weighted bipolar graphs.
\newblock In \emph{ECSQARU 2017}, 25--35. Springer.

\bibitem[{Angwin et~al.(2016)Angwin, Larson, Mattu, and Kirchner}]{compas_paper}
Angwin, J.; Larson, J.; Mattu, S.; and Kirchner, L. 2016.
\newblock Machine Bias: There’s software used across the country to predict future criminals. And it’s biased against blacks.
\newblock \emph{ProPublica}.
\newblock Accessed: 2025-05-06.

\bibitem[{Ayoobi et~al.(2019)Ayoobi, Cao, Verbrugge, and Verheij}]{AyoobiCVV19}
Ayoobi, H.; Cao, M.; Verbrugge, R.; and Verheij, B. 2019.
\newblock Handling Unforeseen Failures Using Argumentation-Based Learning.
\newblock In \emph{CASE 2019}, 1699--1704. {IEEE}.

\bibitem[{Ayoobi et~al.(2021)Ayoobi, Cao, Verbrugge, and Verheij}]{AyoobiCVV21}
Ayoobi, H.; Cao, M.; Verbrugge, R.; and Verheij, B. 2021.
\newblock Argue to Learn: Accelerated Argumentation-Based Learning.
\newblock In Wani, M.~A.; Sethi, I.~K.; Shi, W.; Qu, G.; Raicu, D.~S.; and Jin, R., eds., \emph{{ICMLA} 2021}, 1118--1123. {IEEE}.

\bibitem[{Ayoobi et~al.(2022)Ayoobi, Cao, Verbrugge, and Verheij}]{AyoobiABL}
Ayoobi, H.; Cao, M.; Verbrugge, R.; and Verheij, B. 2022.
\newblock Argumentation-Based Online Incremental Learning.
\newblock \emph{{IEEE} Trans Autom. Sci. Eng.}, 19(4): 3419--3433.

\bibitem[{Ayoobi et~al.(2023{\natexlab{a}})Ayoobi, Kasaei, Cao, Verbrugge, and Verheij}]{ayoobiICRA}
Ayoobi, H.; Kasaei, H.; Cao, M.; Verbrugge, R.; and Verheij, B. 2023{\natexlab{a}}.
\newblock Explain What You See: Open-Ended Segmentation and Recognition of Occluded 3D Objects.
\newblock In \emph{{ICRA} 2023}, 4960--4966. {IEEE}.

\bibitem[{Ayoobi et~al.(2025{\natexlab{a}})Ayoobi, Potyka, Rapberger, and Toni}]{ayoobietal2025}
Ayoobi, H.; Potyka, N.; Rapberger, A.; and Toni, F. 2025{\natexlab{a}}.
\newblock Argumentative Debates for Transparent Bias Detection [Technical Report].
\newblock \emph{CoRR}, abs/2508.04511.

\bibitem[{Ayoobi et~al.(2023{\natexlab{b}})Ayoobi, Potyka, Toni, and and}]{AyoobiSparx}
Ayoobi, H.; Potyka, N.; Toni, F.; and and. 2023{\natexlab{b}}.
\newblock SpArX: Sparse Argumentative Explanations for Neural Networks.
\newblock In \emph{{ECAI} 2023}, 149--156. {IOS} Press.

\bibitem[{Ayoobi et~al.(2025{\natexlab{b}})Ayoobi, Potyka, Toni, and and}]{AyoobiPT25}
Ayoobi, H.; Potyka, N.; Toni, F.; and and. 2025{\natexlab{b}}.
\newblock ProtoArgNet: Interpretable Image Classification with Super-Prototypes and Argumentation.
\newblock In \emph{AAAI-25}, 1791--1799. {AAAI} Press.

\bibitem[{Baroni, Rago, and Toni(2018)}]{BaroniRT18}
Baroni, P.; Rago, A.; and Toni, F. 2018.
\newblock How Many Properties Do We Need for Gradual Argumentation?
\newblock In \emph{AAAI 2018}, 1736--1743. {AAAI} Press.

\bibitem[{Becker and Kohavi(1996)}]{adult_census}
Becker, B.; and Kohavi, R. 1996.
\newblock Adult.
\newblock Accessed: 2025-05-06.

\bibitem[{Brown{-}Cohen, Irving, and Piliouras(2024)}]{debate-safety2024}
Brown{-}Cohen, J.; Irving, G.; and Piliouras, G. 2024.
\newblock Scalable {AI} Safety via Doubly-Efficient Debate.
\newblock In \emph{{ICML} 2024}. OpenReview.net.

\bibitem[{Caton and Haas(2024)}]{survey-bias24}
Caton, S.; and Haas, C. 2024.
\newblock Fairness in Machine Learning: {A} Survey.
\newblock \emph{{ACM} Comput. Surv.}, 56(7): 166:1--166:38.

\bibitem[{Corbett{-}Davies et~al.(2017)Corbett{-}Davies, Pierson, Feller, Goel, and Huq}]{stat-parity17}
Corbett{-}Davies, S.; Pierson, E.; Feller, A.; Goel, S.; and Huq, A. 2017.
\newblock Algorithmic Decision Making and the Cost of Fairness.
\newblock In \emph{{SIGKDD} 2017}, 797--806. {ACM}.

\bibitem[{Cover and Hart(1967)}]{KNN_original_paper}
Cover, T.; and Hart, P. 1967.
\newblock Nearest neighbor pattern classification.
\newblock \emph{IEEE Transactions on Information Theory}, 13(1): 21--27.

\bibitem[{Cyras et~al.(2017)Cyras, Fan, Schulz, and Toni}]{DBLP:journals/flap/CyrasFST17}
Cyras, K.; Fan, X.; Schulz, C.; and Toni, F. 2017.
\newblock Assumption-based Argumentation: Disputes, Explanations, Preferences.
\newblock \emph{{FLAP}}, 4(8).

\bibitem[{de~Tarl{\'{e}}, Bonzon, and Maudet(2022)}]{maudet}
de~Tarl{\'{e}}, L.~D.; Bonzon, E.; and Maudet, N. 2022.
\newblock Multiagent Dynamics of Gradual Argumentation Semantics.
\newblock In \emph{AAMAS 2022}, 363--371. IFAAMAS.

\bibitem[{Dejl et~al.(2025{\natexlab{a}})Dejl, Ayoobi, Cherrington, Gardner, Toni, Pakzad-Shahabi, and Williams}]{Dejl2025}
Dejl, A.; Ayoobi, H.; Cherrington, C.; Gardner, D.; Toni, F.; Pakzad-Shahabi, L.; and Williams, M. 2025{\natexlab{a}}.
\newblock ARGTUMOUR: INTEGRATING LARGE LANGUAGE MODELS AND COMPUTATIONAL ARGUMENTATION TO DISCUSS TREATMENT OPTIONS FOR HIGH-GRADE GLIOMA.
\newblock \emph{Neuro-Oncology}, 27(Supplement\_2): ii26--ii26.

\bibitem[{Dejl et~al.(2025{\natexlab{b}})Dejl, Zhang, Ayoobi, Williams, and Toni}]{AyoobiFacct}
Dejl, A.; Zhang, D.; Ayoobi, H.; Williams, M.; and Toni, F. 2025{\natexlab{b}}.
\newblock Hidden Conflicts in Neural Networks and their Implications for Explainability.
\newblock In \emph{FAccT 2025}, 1498--1542. {ACM}.

\bibitem[{Ehyaei, Farnadi, and Samadi(2024)}]{NEURIPS2024_Fairness}
Ehyaei, A.-R.; Farnadi, G.; and Samadi, S. 2024.
\newblock Wasserstein Distributionally Robust Optimization through the Lens of Structural Causal Models and Individual Fairness.
\newblock In \emph{NeurIPS 2024}, 42430--42467. Curran Associates, Inc.

\bibitem[{Fawkes et~al.(2024)Fawkes, Fishman, Andrews, and Lipton}]{COMPAS_NEURIPS2024}
Fawkes, J.; Fishman, N.; Andrews, M.; and Lipton, Z.~C. 2024.
\newblock The Fragility of Fairness: Causal Sensitivity Analysis for Fair Machine Learning.
\newblock In \emph{NeurIPS 2024}, volume~37, 137105--137134. Curran Associates, Inc.

\bibitem[{Grabowicz, Perello, and Mishra(2022)}]{fairness+explanation22}
Grabowicz, P.~A.; Perello, N.; and Mishra, A. 2022.
\newblock Marrying Fairness and Explainability in Supervised Learning.
\newblock In \emph{FAccT '22}, 1905--1916. {ACM}.

\bibitem[{Irving, Christiano, and Amodei(2018)}]{debate-safety2018}
Irving, G.; Christiano, P.~F.; and Amodei, D. 2018.
\newblock {AI} safety via debate.
\newblock \emph{CoRR}, abs/1805.00899.

\bibitem[{Khademi and Honavar(2020)}]{COMPAS_AAAI2020}
Khademi, A.; and Honavar, V. 2020.
\newblock Algorithmic Bias in Recidivism Prediction: A Causal Perspective (Student Abstract).
\newblock In \emph{AAAI 2020}, 13839--13840.

\bibitem[{Khan et~al.(2024)Khan, Hughes, Valentine, Ruis, Sachan, Radhakrishnan, Grefenstette, Bowman, Rockt{\"{a}}schel, and Perez}]{debateLLMs-2024}
Khan, A.; Hughes, J.; Valentine, D.; Ruis, L.; Sachan, K.; Radhakrishnan, A.; Grefenstette, E.; Bowman, S.~R.; Rockt{\"{a}}schel, T.; and Perez, E. 2024.
\newblock Debating with More Persuasive LLMs Leads to More Truthful Answers.
\newblock In \emph{ICML 2024}. OpenReview.net.

\bibitem[{Kori, Glocker, and Toni(2024)}]{visualdebates24}
Kori, A.; Glocker, B.; and Toni, F. 2024.
\newblock Explaining Image Classifiers with Visual Debates.
\newblock In \emph{DS2024}, volume 15244 of \emph{Lecture Notes in Computer Science}, 200--214. Springer.

\bibitem[{Leite and Martins(2011)}]{LeiteM11}
Leite, J.; and Martins, J.~G. 2011.
\newblock Social Abstract Argumentation.
\newblock In \emph{{IJCAI} 2011}, 2287--2292. {IJCAI/AAAI}.

\bibitem[{Leofante et~al.(2024)Leofante, Ayoobi, Dejl, Freedman, Gorur, Jiang, Paulino{-}Passos, Rago, Rapberger, Russo, Yin, Zhang, and Toni}]{AyoobiKR}
Leofante, F.; Ayoobi, H.; Dejl, A.; Freedman, G.; Gorur, D.; Jiang, J.; Paulino{-}Passos, G.; Rago, A.; Rapberger, A.; Russo, F.; Yin, X.; Zhang, D.; and Toni, F. 2024.
\newblock Contestable {AI} Needs Computational Argumentation.
\newblock In Marquis, P.; Ortiz, M.; and Pagnucco, M., eds., \emph{KR2024}.

\bibitem[{Liu et~al.(2024)Liu, Gautam, Ma, and Lakkaraju}]{LLM_Bias}
Liu, Y.; Gautam, S.; Ma, J.; and Lakkaraju, H. 2024.
\newblock Confronting {LLM}s with Traditional {ML}: Rethinking the Fairness of Large Language Models in Tabular Classifications.
\newblock In Duh, K.; Gomez, H.; and Bethard, S., eds., \emph{NAACL 2024 (Volume 1: Long Papers)}, 3603--3620. Association for Computational Linguistics.

\bibitem[{Ma et~al.(2025)Ma, Salinas, Nyarko, and Henderson}]{BiasLLMBreakingDown}
Ma, S.; Salinas, A.; Nyarko, J.; and Henderson, P. 2025.
\newblock Breaking Down Bias: On The Limits of Generalizable Pruning Strategies.
\newblock In \emph{FAccT '25}, 2437–2450. ACM.

\bibitem[{Macagno, Walton, and Reed(2017)}]{waltonhandbook}
Macagno, F.; Walton, D.; and Reed, C. 2017.
\newblock Argumentation Schemes. History, Classifications, and Computational Applications.
\newblock \emph{{FLAP}}, 4(8).

\bibitem[{Mehrabi et~al.(2022)Mehrabi, Morstatter, Saxena, Lerman, and Galstyan}]{survey-bias21}
Mehrabi, N.; Morstatter, F.; Saxena, N.; Lerman, K.; and Galstyan, A. 2022.
\newblock A Survey on Bias and Fairness in Machine Learning.
\newblock \emph{{ACM} Comput. Surv.}, 54(6): 115:1--115:35.

\bibitem[{Moro, Rita, and Cortez(2014)}]{bank_marketing}
Moro, S.; Rita, P.; and Cortez, P. 2014.
\newblock {Bank Marketing}.
\newblock UCI Machine Learning Repository.
\newblock {DOI}: https://doi.org/10.24432/C5K306.

\bibitem[{Mossakowski and Neuhaus(2018)}]{mossakowski2018modular}
Mossakowski, T.; and Neuhaus, F. 2018.
\newblock Modular semantics and characteristics for bipolar weighted argumentation graphs.
\newblock \emph{arXiv preprint arXiv:1807.06685}.

\bibitem[{Panisson, McBurney, and Bordini(2021)}]{DBLP:journals/argcom/PanissonMB21}
Panisson, A.~R.; McBurney, P.; and Bordini, R.~H. 2021.
\newblock A computational model of argumentation schemes for multi-agent systems.
\newblock \emph{Argument Comput.}, 12(3): 357--395.

\bibitem[{Potyka(2018)}]{potyka2018continuous}
Potyka, N. 2018.
\newblock Continuous dynamical systems for weighted bipolar argumentation.
\newblock In \emph{KR 2018}, 148--157.

\bibitem[{Potyka(2019)}]{Potyka:2019}
Potyka, N. 2019.
\newblock {Extending Modular Semantics for Bipolar Weighted Argumentation}.
\newblock In \emph{AAMAS 2019}, 1722--1730. IFAAMAS.

\bibitem[{Potyka and Booth(2024{\natexlab{a}})}]{PotykaB024}
Potyka, N.; and Booth, R. 2024{\natexlab{a}}.
\newblock Balancing Open-Mindedness and Conservativeness in Quantitative Bipolar Argumentation (and How to Prove Semantical from Functional Properties).
\newblock In \emph{{KR} 2024}, 597--607.

\bibitem[{Potyka and Booth(2024{\natexlab{b}})}]{PotykaB24}
Potyka, N.; and Booth, R. 2024{\natexlab{b}}.
\newblock An Empirical Study of Quantitative Bipolar Argumentation Frameworks for Truth Discovery.
\newblock In \emph{COMMA 2024}, 205--216. {IOS} Press.

\bibitem[{{ProPublica}(2016)}]{compas_kaggle}
{ProPublica}. 2016.
\newblock COMPAS Recidivism Racial Bias Dataset.
\newblock \url{https://www.kaggle.com/datasets/danofer/compass}.
\newblock Accessed: 2025-05-06.

\bibitem[{Rago et~al.(2025)Rago, Cocarascu, Oksanen, and Toni}]{DBLP:journals/ai/RagoCOT25}
Rago, A.; Cocarascu, O.; Oksanen, J.; and Toni, F. 2025.
\newblock Argumentative review aggregation and dialogical explanations.
\newblock \emph{Artif. Intell.}, 340: 104291.

\bibitem[{Rago et~al.(2016)Rago, Toni, Aurisicchio, and Baroni}]{rago2016discontinuity}
Rago, A.; Toni, F.; Aurisicchio, M.; and Baroni, P. 2016.
\newblock Discontinuity-Free Decision Support with Quantitative Argumentation Debates.
\newblock In \emph{KR 2016}, 63--73.

\bibitem[{R\"{a}z(2024)}]{COMPAS_Facct2024}
R\"{a}z, T. 2024.
\newblock Reliability Gaps Between Groups in COMPAS Dataset.
\newblock In \emph{FAccT '24}, 113–126. ACM.

\bibitem[{Thang, Dung, and Hung(2009)}]{DBLP:journals/logcom/ThangDH09}
Thang, P.~M.; Dung, P.~M.; and Hung, N.~D. 2009.
\newblock Towards a Common Framework for Dialectical Proof Procedures in Abstract Argumentation.
\newblock \emph{J. Log. Comput.}, 19(6): 1071--1109.

\bibitem[{Thauvin et~al.(2024)Thauvin, Herbin, Ouerdane, and Hudelot}]{debate4images-ECAI24}
Thauvin, D.; Herbin, S.; Ouerdane, W.; and Hudelot, C. 2024.
\newblock Interpretable Image Classification Through an Argumentative Dialog Between Encoders.
\newblock In \emph{{ECAI} 2024}, 3316--3323. {IOS} Press.

\bibitem[{Waller, Rodrigues, and Cocarascu(2024)}]{OanaBias}
Waller, M.; Rodrigues, O.; and Cocarascu, O. 2024.
\newblock Identifying Reasons for Bias: An Argumentation-Based Approach.
\newblock In \emph{AAAI 2024}, 21664--21672. {AAAI} Press.

\bibitem[{Waller et~al.(2024)Waller, Rodrigues, Lee, and Cocarascu}]{oana-survey24}
Waller, M.; Rodrigues, O.; Lee, M. S.~A.; and Cocarascu, O. 2024.
\newblock Bias Mitigation Methods: Applicability, Legality, and Recommendations for Development.
\newblock \emph{J. Artif. Intell. Res.}, 81: 1043--1078.

\bibitem[{Walton, Reed, and Macagno(2008)}]{waltonbook}
Walton, D.; Reed, C.; and Macagno, F. 2008.
\newblock \emph{Argumentation Schemes}.
\newblock Cambridge University Press.

\bibitem[{Xiang(2024)}]{BiasGenAILaw}
Xiang, A. 2024.
\newblock Fairness \& Privacy in an Age of Generative AI.
\newblock \emph{Science and Technology Law Review}, 25(2).

\end{thebibliography}

\clearpage

\end{document}


\maketitle

\section{Background on QBAF properties}
\label{app:QBAF properties}

We recall some 
notions from \cite{PotykaB24} that we use in this paper. 
Below $A,A',S,S',T$ are multisets of real numbers.
\begin{itemize}
\item 
Let 
$\core(S) = \{x \in S \mid x \neq 0\}$
.   
Then:
\\
$S$ \emph{(weakly) dominates}
$T$ ($S \succeq T$ in short) if $\core(S) = \core(T) = \emptyset$ or there exist 
$S' \subseteq \core(S)$ and a bijective function
$f: \core(T) \rightarrow S'$ such that
$x \leq f(x)$ for all
$x \in \core(T)$.
If, in addition, $|\core(S)| >  |\core(T)|$ or
there is an $x \in \core(T)$ such that $x < f(x)$, $S$ \emph{strictly dominates}
$T$ ($S \succ T$ in short).
\\
$S$ and 
 $T$ are \emph{balanced} ($S \!\cong\! T$) if
$\core(S) \!=\! \core(T)$.
\item An aggregation function 
$\agg$ satisfies 
\emph{balance} iff
\\
 \quad 1. $A \!\cong \! S$ implies $\agg(A ,S) = 0$, and
\\ 
    \quad 2. $A \!\cong\! A'$ and
    $S\! \cong \!S'$ implies $
    \agg(A ,S) \!=\! 
    \agg(A' ,S')$;
\\
\emph{monotonicity} iff
\\
 \quad 1. $\agg(A,S) \!\leq \!0$ if $A \!\succeq \!S$, 
    \\
    \quad 2. $\agg(A,S) \!\geq \! 0$ if $S \!\succeq \! A$, 
   \\ 
    \quad 3. $\agg(A,S) \leq 
    \agg(A',S)$ if $A \succeq A'$, and
    \\
    \quad 4. $\agg(A,S) \geq \agg(A,S')$ if $S \succeq S'$.
\\
If the inequalities can be replaced by strict inequalities
for strict domination, $\agg$ satisfies \emph{strict monotonicity}.

\item An influence function $\infl$ satisfies 
\\
\emph{balance}
iff $\infl(b, \!0) \!\!=\!\! b$;
\\
\emph{monotonicity} iff
\\
\quad     1. $\infl(b, a) \!\leq\! b$ if $a \!< \!0$, 
  %
    \\ \quad
     2. $\infl(b, a) \!\geq\! b$ if $a \!>\! 0$, 
\\    
    \quad 3. $\infl(b_1, a) \leq \infl(b_2, a)$ if $b_1 < b_2$, and
  \\  
    \quad 4. $\infl(b, a_1) \leq \infl(b, a_2)$ if $a_1 < a_2$.
\\
If the inequalities can be replaced by strict 
ones
 when 
 $b \!\neq \! 0$ for 
 1. and 
 $b\!\neq \!1$ for 
 2.,
 $\infl$ satisfies \emph{strict monotonicity}.
\end{itemize}
\section{Proofs}
\label{app:proofs}

Here we give the proofs for all results in the paper.

\paragraph{Proposition~1}

\begin{proof}
To prove the claim, consider two points $\cinput_1, \cinput_2 \in \nb$ such that there is an $\cinput_3 \in S$ that is a convex combination of $\cinput_1$ and $\cinput_2$. Then, we have
$d(\cinput, x_3) = \|\cinput - x_3\| 
=  \|\cinput - (\lambda x_1 + (1-\lambda) x_2)\|
=  \lambda \|\cinput - x_1 \| + (1-\lambda) \| \cinput - x_2\|
\leq \epsilon
$.
\end{proof}

\paragraph{Proof for Proposition~2}

\begin{proof}
1. Since $\advantaged^{att} =  \disadvantaged^{sup}$
and $\advantaged^{sup} =  \disadvantaged^{att}$,
we have $\advantaged^{att} \succeq  \advantaged^{sup}$ iff
$\disadvantaged^{sup} \succeq  \disadvantaged^{att}$.
Hence, by monotonicity, $\sigma(\advantaged)
= \infl(\tau(\advantaged), \agg(\advantaged^{att}, \advantaged^{sup}))
\leq \tau(\advantaged) = 0$ or 
$\sigma(\disadvantaged)
= \infl(\tau(\disadvantaged), \agg(\disadvantaged^{att}, \disadvantaged^{sup}))
\leq \tau(\disadvantaged) = 0$
(or both if $\advantaged^{att} \cong  \disadvantaged^{att}$).

2. The assumptions imply that $\advantaged^{att} \cong \advantaged^{sup}$.
Hence, $\sigma(\advantaged) 
= \infl(\tau(\advantaged), \agg(\advantaged^{att}, \advantaged^{sup}))
= \infl(\tau(\advantaged), 0))
= \tau(\advantaged) = 0$ by balance. The proof for 
$\disadvantaged$ is analogous.

3. The assumptions imply that $\disadvantaged^{sup} \succ \disadvantaged^{att}$.
Hence, $\sigma(\disadvantaged)
=\infl(\tau(\disadvantaged), \agg(\disadvantaged^{att}, \disadvantaged^{sup}))
=\infl(0, \agg(\disadvantaged^{att}, \disadvantaged^{sup}))
> 0$ by strict monotonicity.

4. The proof is analogous to the one for item 3.
\end{proof}

\paragraph{Proof of Proposition~3}

\begin{proof}
1. The assumptions imply that 
$\advantaged^{1, att} = \disadvantaged^{1, sup} \cong \advantaged^{2, att} = \disadvantaged^{2, sup}$ and $\disadvantaged^{1, att} = \advantaged^{1, sup}  \cong \disadvantaged^{2, att} = \advantaged^{2, sup}$.
Hence, balance implies that $\sigma(\advantaged^1) 
= \infl(\tau(\advantaged), \agg(\advantaged^{att}, \advantaged^{sup}))
=\infl(\tau(\disadvantaged), \agg(\disadvantaged^{att}, \disadvantaged^{sup}))
= \sigma(\advantaged^2)$ and $\sigma(\disadvantaged^1) = \sigma(\disadvantaged^2)$. 

2. The assumptions imply that 
$\advantaged^{1, att} = \disadvantaged^{1, sup} \succeq \advantaged^{2, att} = \disadvantaged^{2, sup}$ and $\disadvantaged^{2, att} = \advantaged^{2, sup} \succeq \disadvantaged^{1, att} = \advantaged^{1, sup}$.
Hence, monotonicity implies that $\sigma(\advantaged^1) \leq \sigma(\advantaged^2)$ and $\sigma(\disadvantaged^1) \geq \sigma(\disadvantaged^2)$. 

3. The proof is similar to the previous one, just using the additional strictness assumptions and strict monotonicity instead of monotonicity.

Items 4 and 5 follow symmetrically to items 2 and 3.
\end{proof}

\paragraph{Proof of Proposition~4}

\begin{proof}
By assumption, we have
$\sigma(\advantaged^1) 
= \infl(0, \agg(\{P_{\nb^1}(c(\bX)=1 \mid X_p \neq g)\} ,\{P_{\nb^1}(c(\bX)=1 \mid X_p = g)\}))
= \infl(0, \agg(\{P_{\nb^2}(c(\bX)=1 \mid X_p \neq g)\}, \{P_{\nb^2}(c(\bX)=1 \mid X_p = g)\}))
= \sigma(\disadvantaged^2). 
$  
$\sigma(\disadvantaged^1) = \sigma(\advantaged^2)$ follows analogously.
\end{proof}

\paragraph{Proof of Proposition~5}

\begin{proof}
1. For DF-QuAD, we have
$
 \sigma(\advantaged) 
= \infl(0, \agg(\{P_{\nb^1}(c(\bX)=1 \mid X_p \neq g)\} ,\{P_{\nb^1}(c(\bX)=1 \mid X_p = g)\})) 
= \infl(0, (1 - P_{\nb^1}(c(\bX)=1 \mid X_p \neq g)\}) - (1 - P_{\nb^1}(c(\bX)=1 \mid X_p = g))) 
= \infl(0, P_{\nb^1}(c(\bX)=1 \mid X_p = g) - P_{\nb^1}(c(\bX)=1 \mid X_p \neq g)) 
= \max \{ 0, P_{\nb^1}(c(\bX)=1 \mid X_p = g) - P_{\nb^1}(c(\bX)=1 \mid X_p \neq g)\}.   
$

The proof of item 2 is symmetrical. Item 3 follows directly from item 2 by definition of $\epsilon$-biasedness.
\end{proof}

\paragraph{Proof of Proposition~6}

\begin{proof}
 By assumption, we have
$\sigma(\topicarg) 
= \infl(\bases(\topicarg), \agg(\topicarg^{att} , \topicarg^{sup}))
= \infl(0, \agg(S_a, S_d)), 
$     
where we used balance of $\agg$ for the second inequality. Now claim 1 follows immediately from balance
and claim 2 from monotonicity of $\agg$ and $\infl$.
\end{proof}

\section{Experimental details}
\label{app:experiments}

\subsection{Experimental set-up}
In all experiments, we have used $\epsilon = 0.01$ and $N = 30$.  
We also tested values of \(\epsilon \in \{0.001, 0.005\}\)
and $N \in \{10,20\}$ all yielding similar results. The experiments do not use any GPU, and they are fully executable on a single CPU.

\subsection{
Synthetically Biased Models}\label{app:synthetic_models}
\begin{figure}[t]
    \centering
    \includegraphics[width=1\linewidth]{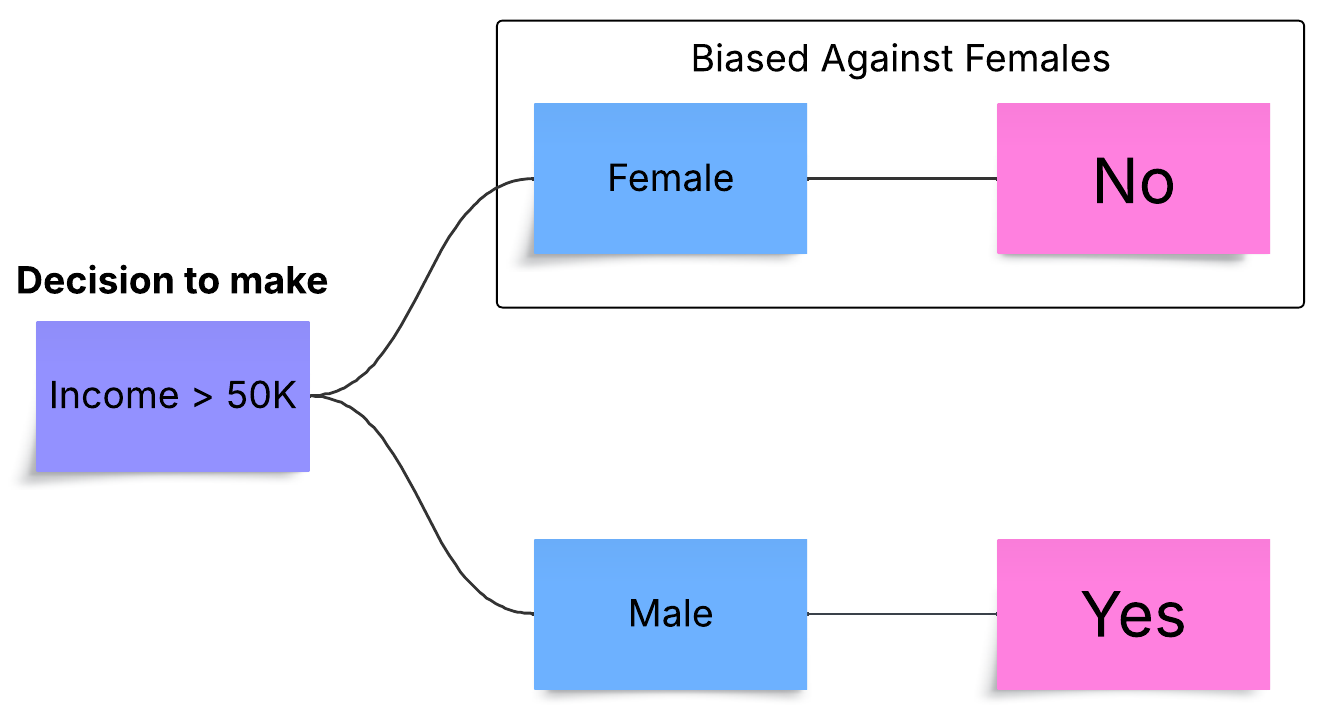}
    \caption{Decision tree showing the reason for global bias against females in the first global synthetic model (Global 1).}
    \label{fig:dt-global-1}
\end{figure}

\begin{figure}
    \centering
    \includegraphics[width=1\linewidth]{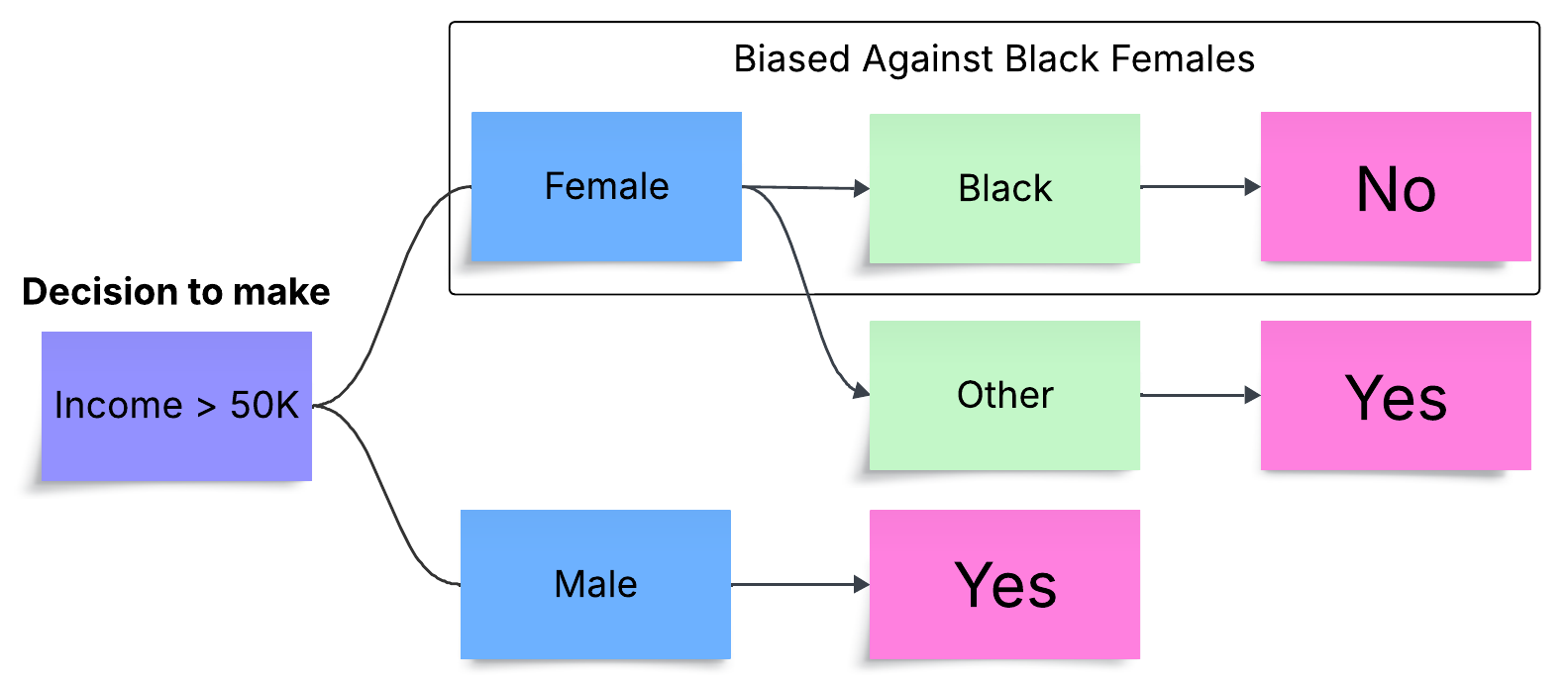}
    \caption{Decision tree showing the reason for global bias against black females in the second global synthetic model (Global 2).}
    \label{fig:dt-global-2}
\end{figure}

\begin{figure}
    \centering
    \includegraphics[width=1\linewidth]{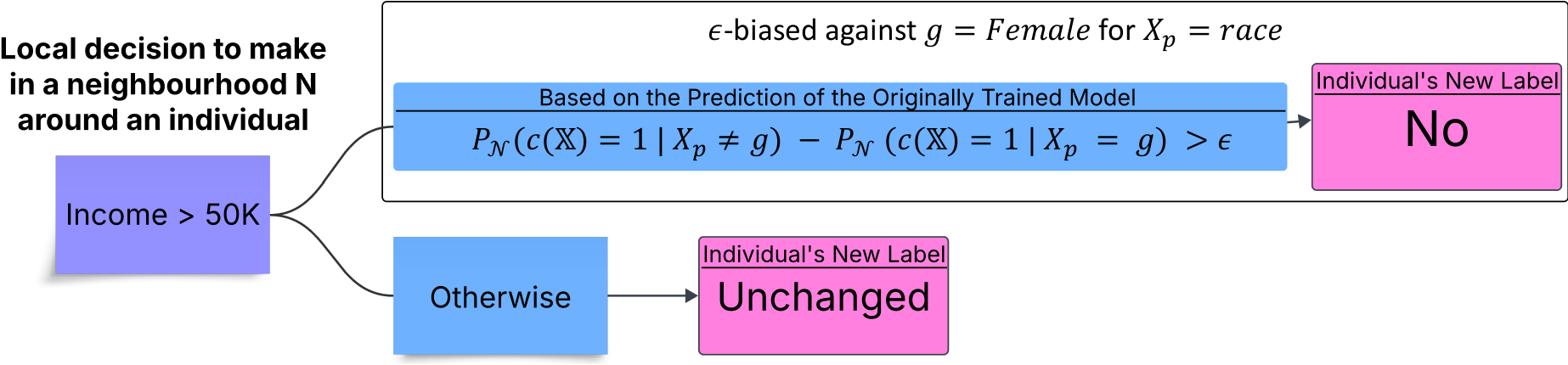}
    \caption{Decision tree showing the reason for local bias against females in a neighbourhood around an individual in the local synthetic model (Local 1).}
    \label{fig:dt-local-1}
\end{figure}

In this section, we have visualized the decision trees for the systematically biased models (in Section 7). Figure \ref{fig:dt-global-1} shows a decision tree that is globally biased against females (Global 1). Figure \ref{fig:dt-global-2} illustrates the second globally biased model (Global 2), which is biased against all black females. For the locally biased model (Local 1), we first trained a logistic regression model on the 
ACI training set. In contrast to the global synthetically biased  models, Local 1 operates at a neighbourhood level, changing the
prediction to the negative label for female individuals if their neighbourhood exhibits an $\epsilon$-bias against $gender=female$ in the originally trained model (as shown in Figure \ref{fig:dt-local-1}). 

Moreover, for the detection of bias in single neighbourhoods, Table \ref{tab:extended_single_neighbourhood_variance} shows the comparison of our model to IRB with various neighbourhood sizes from 10 to 200 with an interval of 10.

\begin{table}
\centering
\resizebox{\linewidth}{!}{
\begin{tabular}{lcccccc}
\toprule
Method  & K & Accuracy & Precision & Recall & F1-score & Runtime (s) \\
\midrule
\multicolumn{7}{c}{\textbf{Global 1}} \\
\midrule
Our & \multirow{2}{*}{10} & 0.90 ± 0.01 & 0.96 ± 0.01 & 0.87 ± 0.02 & 0.91 ± 0.01 & \textbf{3.84 ± 0.03} \\
IRB &  & 0.90 ± 0.01 & 0.96 ± 0.01 & 0.87 ± 0.01 & 0.91 ± 0.02 & 5.80 ± 0.04 \\
Our & \multirow{2}{*}{20} & 0.92 ± 0.01 & 0.97 ± 0.01 & 0.88 ± 0.02 & 0.92 ± 0.01 & \textbf{3.85 ± 0.02} \\
IRB &  & 0.92 ± 0.01 & 0.97 ± 0.01 & 0.88 ± 0.01 & 0.92 ± 0.02 & 11.16 ± 0.04 \\
Our & \multirow{2}{*}{30} & 0.94 ± 0.01 & 0.98 ± 0.01 & 0.90 ± 0.02 & 0.94 ± 0.01 & \textbf{3.86 ± 0.03} \\
IRB &  & 0.94 ± 0.01 & 0.98 ± 0.01 & 0.90 ± 0.01 & 0.94 ± 0.02 & 16.51 ± 0.05 \\
Our & \multirow{2}{*}{40} & 0.95 ± 0.01 & 0.99 ± 0.00 & 0.90 ± 0.02 & 0.95 ± 0.01 & \textbf{3.87 ± 0.02} \\
IRB &  & 0.95 ± 0.01 & 0.99 ± 0.00 & 0.90 ± 0.01 & 0.95 ± 0.02 & 21.87 ± 0.05 \\
Our & \multirow{2}{*}{50} & 0.95 ± 0.01 & 1.00 ± 0.00 & 0.90 ± 0.02 & 0.95 ± 0.01 & \textbf{3.87 ± 0.03} \\
IRB &  & 0.95 ± 0.01 & 1.00 ± 0.00 & 0.90 ± 0.01 & 0.95 ± 0.02 & 28.22 ± 0.04 \\
Our & \multirow{2}{*}{60} & 0.96 ± 0.01 & 1.00 ± 0.00 & 0.92 ± 0.01 & 0.95 ± 0.01 & \textbf{3.86 ± 0.02} \\
IRB &  & 0.96 ± 0.01 & 1.00 ± 0.00 & 0.92 ± 0.01 & 0.95 ± 0.01 & 32.68 ± 0.04 \\
Our & \multirow{2}{*}{80} & 0.98 ± 0.00 & 1.00 ± 0.00 & 0.96 ± 0.01 & 0.98 ± 0.00 & \textbf{3.87 ± 0.02} \\
IRB &  & 0.98 ± 0.00 & 1.00 ± 0.00 & 0.96 ± 0.01 & 0.98 ± 0.00 & 39.82 ± 0.03 \\
Our & \multirow{2}{*}{90} & 0.99 ± 0.00 & 1.00 ± 0.00 & 0.98 ± 0.01 & 0.99 ± 0.00 & \textbf{3.84 ± 0.02} \\
IRB &  & 0.99 ± 0.00 & 1.00 ± 0.00 & 0.98 ± 0.01 & 0.99 ± 0.00 & 44.10 ± 0.03 \\
Our & \multirow{2}{*}{100} & 1.00 ± 0.00 & 1.00 ± 0.00 & 1.00 ± 0.00 & 1.00 ± 0.00 & \textbf{3.88 ± 0.02} \\
IRB &  & 1.00 ± 0.00 & 1.00 ± 0.00 & 1.00 ± 0.00 & 1.00 ± 0.00 & 48.47 ± 0.05 \\
Our & \multirow{2}{*}{110--200} & 1.00 ± 0.00 & 1.00 ± 0.00 & 1.00 ± 0.00 & 1.00 ± 0.00 & \textbf{3.81--3.87} \\
IRB &  & 1.00 ± 0.00 & 1.00 ± 0.00 & 1.00 ± 0.00 & 1.00 ± 0.00 & 52.83--97.80 \\
\midrule
\multicolumn{7}{c}{\textbf{Global 2}} \\
\midrule
Our & 10 & \textbf{0.91 ± 0.01} & \textbf{0.92 ± 0.01} & \textbf{0.91 ± 0.02} & \textbf{0.91 ± 0.01} & \textbf{0.32 ± 0.02} \\
Our & 20 & \textbf{0.93 ± 0.01} & \textbf{0.95 ± 0.01} & \textbf{0.93 ± 0.02} & \textbf{0.94 ± 0.01} & \textbf{0.32 ± 0.01} \\
Our & 30 & \textbf{0.95 ± 0.01} & \textbf{0.97 ± 0.01} & \textbf{0.95 ± 0.01} & \textbf{0.96 ± 0.01} & \textbf{0.33 ± 0.02} \\
Our & 40 & \textbf{0.96 ± 0.02} & \textbf{0.99 ± 0.00} & \textbf{0.96 ± 0.02} & \textbf{0.98 ± 0.01} & \textbf{0.33 ± 0.02} \\
Our & 50 & \textbf{0.96 ± 0.02} & \textbf{1.00 ± 0.00} & \textbf{0.97 ± 0.01} & \textbf{0.98 ± 0.01} & \textbf{0.34 ± 0.02} \\
Our & 60 & \textbf{0.96 ± 0.02} & \textbf{1.00 ± 0.00} & \textbf{0.96 ± 0.02} & \textbf{0.98 ± 0.01} & \textbf{0.34 ± 0.02} \\
Our & 70 & \textbf{0.97 ± 0.01} & \textbf{1.00 ± 0.00} & \textbf{0.97 ± 0.02} & \textbf{0.98 ± 0.01} & \textbf{0.34 ± 0.01} \\
Our & 80 & \textbf{0.98 ± 0.01} & \textbf{1.00 ± 0.00} & \textbf{0.96 ± 0.01} & \textbf{0.98 ± 0.01} & \textbf{0.34 ± 0.01} \\
Our & 90 & \textbf{0.99 ± 0.01} & \textbf{1.00 ± 0.00} & \textbf{0.98 ± 0.02} & \textbf{0.99 ± 0.01} & \textbf{0.35 ± 0.01} \\
Our & 100--200 & \textbf{1.00 ± 0.00} & \textbf{1.00 ± 0.00} & \textbf{1.00 ± 0.00} & \textbf{1.00 ± 0.00} & \textbf{0.35--0.38 } \\
IRB & 10--200 & 0.00 ± 0.00 & 0.00 ± 0.00 & 0.00 ± 0.00 & 0.00 ± 0.00 & 0.86--17.44 \\
\midrule
\multicolumn{7}{c}{\textbf{Local 1}} \\
\midrule
Our & 10--200 & \textbf{1.00 ± 0.00} & \textbf{1.00 ± 0.00} & \textbf{1.00 ± 0.00} & \textbf{1.00 ± 0.00} & \textbf{5.00 ± 0.05} \\
IRB & 10 & 0.65 ± 0.02 & 0.76 ± 0.01 & 0.36 ± 0.02 & 0.49 ± 0.02 & 10.20 ± 0.03 \\
IRB & 20 & 0.68 ± 0.01 & 0.78 ± 0.01 & 0.38 ± 0.01 & 0.51 ± 0.01 & 15.33 ± 0.04 \\
IRB & 30 & 0.83 ± 0.01 & \textbf{1.00 ± 0.00} & 0.46 ± 0.02 & 0.63 ± 0.01 & 22.12 ± 0.05 \\
IRB & 40 & 0.80 ± 0.01 & \textbf{1.00 ± 0.00} & 0.46 ± 0.02 & 0.63 ± 0.01 & 28.42 ± 0.05 \\
IRB & 50 & 0.81 ± 0.02 & \textbf{1.00 ± 0.00} & 0.44 ± 0.02 & 0.61 ± 0.02 & 30.66 ± 0.05 \\
IRB & 60 & 0.78 ± 0.02 & \textbf{1.00 ± 0.00} & 0.38 ± 0.02 & 0.55 ± 0.02 & 35.82 ± 0.04 \\
IRB & 70 & 0.76 ± 0.02 & \textbf{1.00 ± 0.00} & 0.35 ± 0.02 & 0.52 ± 0.02 & 38.60 ± 0.04 \\
IRB & 80 & 0.75 ± 0.01 & \textbf{1.00 ± 0.00} & 0.33 ± 0.02 & 0.50 ± 0.02 & 42.50 ± 0.05 \\
IRB & 90 & 0.75 ± 0.01 & \textbf{1.00 ± 0.00} & 0.32 ± 0.02 & 0.48 ± 0.02 & 46.20 ± 0.05 \\
IRB & 100 & 0.74 ± 0.02 & \textbf{1.00 ± 0.00} & 0.31 ± 0.02 & 0.48 ± 0.02 & 49.45 ± 0.05 \\
IRB & 110 & 0.73 ± 0.02 & \textbf{1.00 ± 0.00} & 0.29 ± 0.02 & 0.45 ± 0.02 & 52.80 ± 0.05 \\
IRB & 120 & 0.72 ± 0.02 & \textbf{1.00 ± 0.00} & 0.27 ± 0.02 & 0.43 ± 0.02 & 58.00 ± 0.05 \\
IRB & 130 & 0.71 ± 0.02 & \textbf{1.00 ± 0.00} & 0.25 ± 0.02 & 0.40 ± 0.02 & 62.50 ± 0.05 \\
IRB & 140 & 0.71 ± 0.02 & \textbf{1.00 ± 0.00} & 0.24 ± 0.02 & 0.39 ± 0.02 & 67.20 ± 0.05 \\
IRB & 150 & 0.71 ± 0.02 & \textbf{1.00 ± 0.00} & 0.23 ± 0.02 & 0.37 ± 0.02 & 72.10 ± 0.05 \\
IRB & 160 & 0.71 ± 0.01 & \textbf{1.00 ± 0.00} & 0.22 ± 0.01 & 0.36 ± 0.01 & 78.00 ± 0.04 \\
IRB & 170 & 0.70 ± 0.01 & \textbf{1.00 ± 0.00} & 0.22 ± 0.01 & 0.36 ± 0.01 & 84.00 ± 0.04 \\
IRB & 180 & 0.70 ± 0.01 & \textbf{1.00 ± 0.00} & 0.22 ± 0.01 & 0.36 ± 0.01 & 90.00 ± 0.04 \\
IRB & 190 & 0.70 ± 0.01 & \textbf{1.00 ± 0.00} & 0.22 ± 0.01 & 0.36 ± 0.01 & 95.00 ± 0.04 \\
IRB & 200 & 0.70 ± 0.01 & \textbf{1.00 ± 0.00} & 0.22 ± 0.01 & 0.36 ± 0.01 & 97.16 ± 0.04 \\
\bottomrule
\end{tabular}
}
\caption{Comparison of our method versus IRB across multiple neighbourhood sizes ($K$). Best results shown in bold.}
\label{tab:extended_single_neighbourhood_variance}
\end{table}



\subsection{
 Experiments with the Trained Models}

To further explore the findings presented in Table 3.a, we have included additional results. Table \ref{tab:compas_biased_examples} showcases several randomly selected examples where our method identifies biases against African-American individuals. These cases are supported by significantly lower conditional probabilities of positive outcomes for this group, whereas the IRB method often does not flag them as biased. Conversely, there are instances where IRB flags certain cases involving Hispanic and Other groups as biased, but our approach does not, aligning with the observed conditional probabilities. 

\begin{table}[h]
\centering
\resizebox{\linewidth}{!}{%
\begin{tabular}{|c|c|c|c|c|c|c|}
\hline
 \multicolumn{2}{|c|}{\textbf{Protected Feature ($X_p$)}} & \bm{$P_{\nb}(C=1\,|\,X_p=1)$} & \bm{$P_{\nb}(C=1\,|\,X_p=0)$} & \textbf{Our} & \textbf{IRB} \\
\hline
\multirow{6}{*}{\rotatebox[origin=c]{90}{race}} & African-American & 0.47 & 0.85 & Biased & Not Biased \\
 &African-American & 0.40 & 0.71 & Biased & Not Biased \\ 
 &Hispanic         & 0.60 & 0.34 & Not Biased & Biased \\
 &Hispanic         & 0.61 & 0.59 & Not Biased & Biased \\ 
 &Other            & 0.61 & 0.54 & Not Biased & Biased \\ 
 &Other            & 0.60 & 0.55 & Not Biased & Biased \\ \hline
 \rotatebox[origin=c]{90}{sex}&Female & 0.84 & 0.83 & Not Biased & Biased \\
\hline
\end{tabular}
}
\caption{Randomly selected examples from the COMPAS dataset: each row shows a different, negatively classified individual, its protected feature, local conditional probabilities from Definition 7  for 
$|\nb|=K=200$.}
\label{tab:compas_biased_examples}
\end{table}

To further explore the findings presented in Table 3.b, we have included additional results. Table \ref{tab:bank_biased_examples} showcases 
some additional randomly selected examples from these groups 
showing 
that our approach is capable of uncovering meaningful instances of unfairness. Conversely, for the ``Divorced'' category, where IRB has more biased cases based on the distribution of protected features in the neighbourhood 
around the individuals,  it is appropriate not to consider 
them as biased against.

\begin{table}[h]
\centering
\resizebox{\linewidth}{!}{%
\begin{tabular}{|c|c|c|c|c|c|c|}
\hline
 \multicolumn{2}{|c|}{\textbf{Protected Feature ($X_p$)}} & \bm{$P_{\nb}(C=1\,|\,X_p=1
 $} & \bm{$P_{\nb}(C=1\,|\,X_p=0
 $} & \textbf{Our} & \textbf{IRB} \\
\hline
 \multirow{2}{*}{\rotatebox[origin=c]{90}{age}}& YoungOrOld & 0.53 & 0.88 & Biased & Not Biased \\
 &YoungOrOld & 0.60 & 0.91 & Biased & Not Biased \\ \hline
\multirow{5}{*}{\rotatebox[origin=c]{90}{marital}}&Married         & 0.48 & 0.94 & Biased & Not Biased \\
 &Married         & 0.70 & 0.99 & Biased & Not Biased \\ 
  &Single     & 0.60 & 0.77 & Biased & Not Biased \\
 &Divorced            & 0.81 & 0.80 & Not Biased & Biased \\ 
&Divorced            & 0.91 & 0.88 & Not Biased & Biased \\ 
\hline
\end{tabular}
}
\caption{Examples from Bank Marketing
: each row shows a different, negatively classified individual, for $|\nb|\!=\!K\!=\!200$.  
}
\label{tab:bank_biased_examples}
\end{table}


\subsection{Sensitivity Analysis}
\subsubsection{Global Analysis}
We conducted a global, non-parametric sensitivity analysis using the Friedman test following \cite{Saltelli2008}. For each parameter, F1-score distributions across its tested levels were compared while all other parameters varied, and results were averaged across the three synthetic models. The evaluated parameters were: $K$ (here noted as K\_set\_str for bias detection across neighbourhoods) $\in$ \{[25, 50], [25, 50, 100], [25, 50, 100, 200]\}; Epsilon $\in$ \{0.01, 0.05, 0.1\}; Diversity\_Threshold $\in$ \{0.1, 0.25, 0.5\}; Significance\_Threshold $\in$ \{10, 25, 50, 100\}; and Distance $\in$ \{Hamming, Euclidean, Manhattan\}. Parameters with $p < 0.05$ were labeled as sensitive. The model is sensitive to K, Epsilon, Diversity\_Threshold, and Significance\_Threshold, with the latter showing the strongest effect. Distance shows no detectable influence, indicating the model is not sensitive to this parameter within the explored range. 
\begin{table}[t]
\centering
\resizebox{\linewidth}{!}{
\begin{tabular}{lrrr}
\hline
\textbf{Parameter} & \textbf{Chi2} & \textbf{$p$-value} & \textbf{Effect ($p < 0.5$)} \\
\hline
K              & 6.00000  & 0.0497871 & Sensitive \\
Distance                 & 0.80000  & 0.6703200 & Not Sensitive \\
Epsilon                  & 6.00000  & 0.0497871 & Sensitive \\
Diversity\_Threshold     & 6.00000  & 0.0497871 & Sensitive \\
Significance\_Threshold  & 8.11111  & 0.0437703 & Sensitive (Strong) \\
\hline
\end{tabular}}

\caption{Friedman-test sensitivity analysis across parameter levels.}
\end{table}
\subsubsection{OAT Analysis}
To further investigate the influence of individual hyperparameters on model performance, we performed a one-parameter-at-a-time (OAT) sensitivity analysis, averaged across the three synthetic models. In this procedure, all parameters except the one under study were fixed at their best-performing values (determined as the combination achieving the highest mean F1-score across models). For each parameter, the mean and standard deviation of the F1-score were computed across all levels of that parameter while the others were held constant.

\begin{figure}[h!]
    \centering
    \resizebox{\linewidth}{!}{%
    \includegraphics[width=0.7\textwidth]{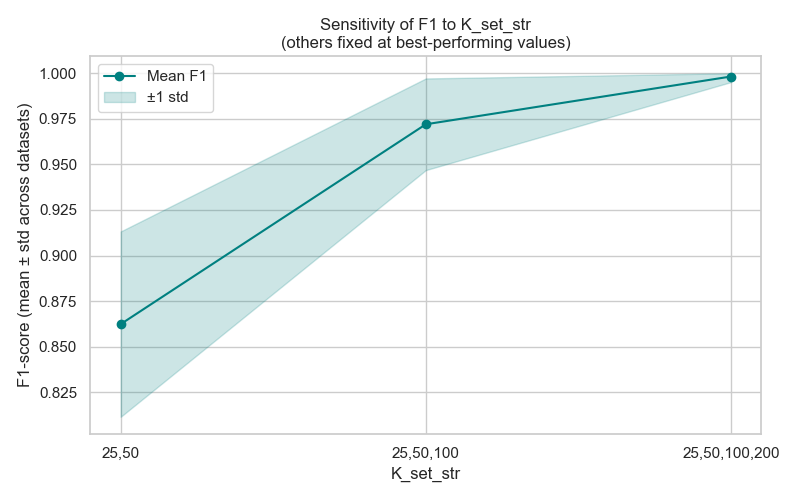}}
    \caption{Sensitivity of F1-score to \texttt{K} (\texttt{K\_set\_str}) with other parameters fixed at best values, averaged across synthetic models.}
    \label{fig:F1_sensitivity_K_set_str}
    
\end{figure}

\begin{figure}[h!]
    \centering
    \resizebox{\linewidth}{!}{%
    \includegraphics[width=0.7\textwidth]{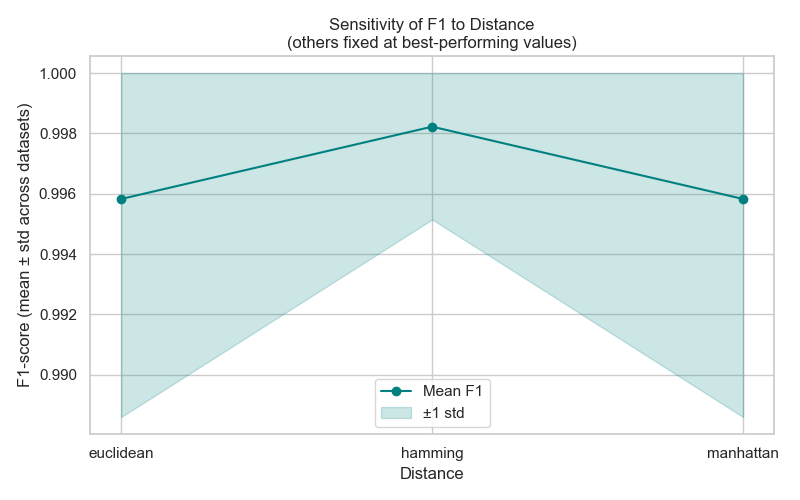}}
    \caption{Sensitivity of F1-score to the \texttt{Distance metric} with other parameters fixed at best values, averaged across synthetic models.}
    \label{fig:F1_sensitivity_distance}
\end{figure}

\begin{figure}[h!]
    \centering
    \resizebox{\linewidth}{!}{%
    \includegraphics[width=0.7\textwidth]{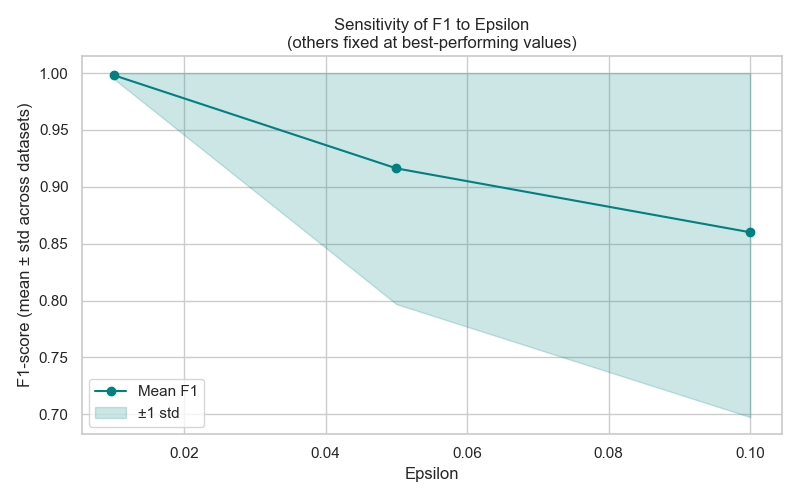}}
    \caption{Sensitivity of F1-score to the \texttt{Epsilon} with other parameters fixed at best values, averaged across synthetic models.}
    \label{fig:F1_sensitivity_epsilon}
\end{figure}

\begin{figure}[h!]
    \centering
    \resizebox{\linewidth}{!}{%
    \includegraphics[width=0.7\textwidth]{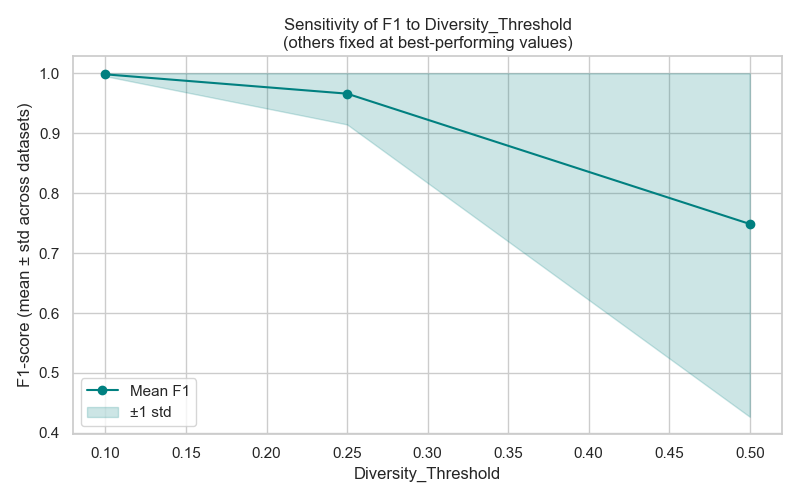}}
    \caption{Sensitivity of F1-score to the \texttt{Diversity Threshold} with other parameters fixed at best values, averaged across synthetic models.}
    \label{fig:F1_sensitivity_diversity}
\end{figure}

\begin{figure}[h!]
    \centering
    \resizebox{\linewidth}{!}{%
    \includegraphics[width=0.7\textwidth]{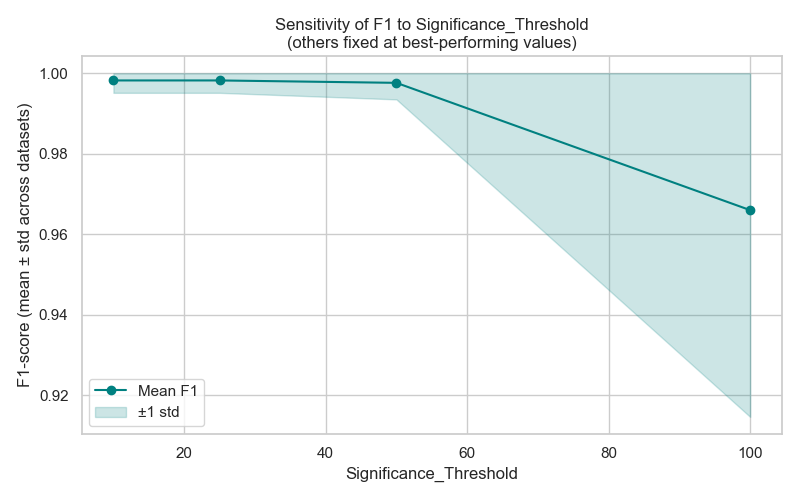}}
    \caption{Sensitivity of F1-score to the \texttt{Significance Threshold} with other parameters fixed at best values, averaged across synthetic models.}
    \label{fig:F1_sensitivity_significance}
\end{figure}

Figures~\ref{fig:F1_sensitivity_K_set_str}--\ref{fig:F1_sensitivity_significance} illustrate the results. Each plot shows the mean F1-score (line) with a shaded region representing $\pm$1 standard deviation across datasets. These visualizations allow us to assess the relative sensitivity of the model to changes in each parameter around the optimal configuration. 

The analysis reveals varying degrees of sensitivity, not just in mean performance but also in cross-dataset stability, as indicated by the $\pm$1 standard deviation shaded regions. \textbf{Distance} clearly has a minimal impact; the mean F1-score remains high (0.996--0.998), and the \textbf{shaded region is consistently narrow}, indicating stable, high performance regardless of the metric used.

In contrast, other parameters show significant sensitivity in both mean and variance:
\begin{itemize}
    \item \textbf{K} (here noted as K\_set\_str) exhibits a strong positive correlation, with the mean F1-score climbing from $\sim$0.86 to nearly 1.0. In particular, the shaded region is wide at the lower-performing ``25,50'' setting and narrows as performance improves, suggesting that larger sets lead to more consistent, high performance.
    \item \textbf{Epsilon} shows a clear negative trend, with performance dropping from 1.0 to $\sim$0.86. This drop in mean performance is accompanied by a progressive widening of the shaded region, signifying that higher Epsilon values not only reduce performance but also make it more inconsistent across the synthetic models.
    \item \textbf{Significance\_Threshold} and Diversity\_Threshold demonstrate threshold effects. Significance\_Threshold is stable (high mean F1, narrow variance) until its value exceeds 50, after which the F1-score drops and the variance increases. The effect is most dramatic for Diversity\_Threshold: after 0.25, the mean F1 drops to $\sim$0.75 and the shaded region expands significantly, indicating unstable and dataset-dependent results in this setting.
\end{itemize}

These OAT results complement the global Friedman-test analysis by highlighting this specific local behavior and cross-dataset (in)stability around the optimal hyperparameter configuration.


\section{Symbols and Notation}
Table \ref{tab:glossary} provides a glossary of the most important symbols and notation used in the paper.

\begin{table}[t]
\centering
\caption{Glossary of Symbols and Notation}
\resizebox{\columnwidth}{!}{
\rowcolors{2}{gray!10}{white}
\begin{tabular}{@{}lp{7cm}@{}}
\toprule
\textbf{Symbol} & \textbf{Meaning} \\
\midrule
\bX & Variable vector $\bX = (X_1, \dots, X_k)$ \\
$X_i$ & $i$-th variable/feature \\
$D_i$ & Domain of feature $X_i$ \\
$\bV$ & Vector of subset of variables \\
$\bv$ & Assignment to $\bV$ \\
\dom & Input space: $\dom = \times_{i=1}^k D_i$ \\
\bx & An instance/input point in $\dom$ \\
$\bx|\bV$ & Projection of $\bx$ on subset of features $\bV$ \\
\coutput & Class label \\
\classes & Set of class labels ($\{0,1\}$) \\
\classifier & Classifier mapping $\bx$ to label: $\classifier: \dom \rightarrow \classes$ \\
$X_p$ & Protected attribute (e.g., gender, race) \\
$g$ & Protected group value
\\
$\nb$ & Neighbourhood of individuals $\nb \subseteq \dom$ \\
$|\nb|$ & Cardinality (number of elements) in neighbourhood \\
$P_\nb(\bv)$ & Local probability of partial assignment $\bv$ in neighbourhood $\nb$ \\
$P_\nb(\classifier(\bX) = 1 \mid \bv)$ & local success probability for $\bv$ in $\nb$ \\
$H_\nb(X_p)$ & entropy of the distribution of the protected feature \\
$\qbaf = (A, R^-, R^+, \tau)$ & QBAF with arguments $A$, attack relation $R^-$,  support relation $R^+$, base score function $\tau$ \\
$a^{\mathit{att}}, a^{\mathit{sup}}$ & Attackers and supporters of argument $a$ \\
$\strength$ & Strength function for QBAF $\qbaf$ \\
$\agg$ & Aggregation function combining attacker/supporter influence  \\
$\infl$ & Influence function modifying base score based on $\agg$ \\
$\succ$ & Strict dominance relation \\
$\succeq$ & Weak dominance relation \\
$\sim$ & Balance relation  \\
\disadvantaged & Argument stating group $g$ is disadvantaged \\
\advantaged & Argument stating group $g$ is advantaged \\
\posG & Argument stating protected group receives positive outcome \\
\posNotG & Argument stating complement of protected group receives positive outcome \\
$s, o, d$ & Critical question arguments for significance, objectivity, diversity \\
$f_s, f_o, f_d$ & Base score defining functions for $s$, $o$, $d$ respectively \\
$\singlebaf_\nb$ & Local bias-QBAF for neighbourhood $\nb$ \\
$\singlebaf^c_\nb$ & Local bias-QBAF with critical questions \\
$\metaArgu_g$ & Global bias-QBAF constructed from local QBAFs \\
$\mathit{bias}_g$ & Global argument stating that classifier is biased against protected group \\
\bottomrule
\end{tabular}
}
\label{tab:glossary}
\end{table}

\bibliography{aaai2026}